\documentclass[acmtog]{acmart}
\acmSubmissionID{561}

\usepackage{multirow}
\usepackage{amsmath}
\usepackage{xcolor}
\usepackage{booktabs} %
\usepackage{acronym} %
\usepackage[normalem]{ulem} 
\usepackage{balance}
\usepackage{hyperref}
\hypersetup{
    colorlinks=true,
    linkcolor=blue,
    filecolor=magenta,      
    urlcolor=cyan,
    pdftitle={Overleaf Example},
    pdfpagemode=FullScreen,
    }
\usepackage[ruled]{algorithm2e} %

\SetAlFnt{\small}
\SetAlCapFnt{\small}
\SetAlCapNameFnt{\small}
\SetAlCapHSkip{0pt}

\copyrightyear{2022}
\acmYear{2022}
\setcopyright{rightsretained}
\acmConference[SA '22 Conference Papers]{SIGGRAPH Asia 2022 Conference Papers}{December 6--9, 2022}{Daegu, Republic of Korea}
\acmBooktitle{SIGGRAPH Asia 2022 Conference Papers (SA '22 Conference Papers), December 6--9, 2022, Daegu, Republic of Korea}
\acmDOI{10.1145/3550469.3555423}
\acmISBN{978-1-4503-9470-3/22/12}

\newcommand{\modelname}{SCARF\xspace}
\newcommand{\modelnamelong}{Segmented Clothed Avatar Radiance Field\xspace}

\newcommand{\qheading}[1]{\noindent\textbf{#1}. }
\newcommand{\supmat}{Sup.~Mat.\xspace}
\newcommand{\smplx}{\mbox{SMPL-X}\xspace}

\acrodef{NeRF}{neural radiance fields}

\newcommand{\mpi}{Max Planck Institute for Intelligent Systems}

\begin{document}
\title{Capturing and Animation of Body and Clothing from Monocular Video}

\author{Yao Feng}
\affiliation{%
\institution{\mpi~}
\city{T{\"u}bingen}
\country{Germany}}
\affiliation{%
\institution{ETH Z{\"u}rich}
\city{Z{\"u}rich}
\country{Switzerland}}
\email{yfeng@tuebingen.mpg.de}

\author{Jinlong Yang}
\affiliation{%
\institution{\mpi}
\city{T{\"u}bingen}
\country{Germany}}
\email{jyang@tuebingen.mpg.de}

\author{Marc Pollefeys}
\affiliation{%
\institution{ETH Z{\"u}rich}
\city{Z{\"u}rich}
\country{Switzerland}}
\email{marc.pollefeys@inf.ethz.ch}

\author{Michael J. Black}
\affiliation{%
\institution{\mpi}
\city{T{\"u}bingen}
\country{Germany}}
\email{black@tuebingen.mpg.de}

\author{Timo Bolkart}
\affiliation{%
\institution{\mpi}
\city{T{\"u}bingen}
\country{Germany}}
\email{tbolkart@tuebingen.mpg.de}

\newcommand{\vect}[1]{\mathbf{#1}}
\newcommand{\norm}[1]{\left\lVert#1\right\rVert}

\newcommand{\shapecoeff}{\boldsymbol{\beta}}
\newcommand{\shapedim}{{\left| \shapecoeff \right|}}
\newcommand{\shapespace}{\mathcal{S}}
\newcommand{\shapespaceexpl}{\mathbb{R}^{\shapedim}}

\newcommand{\numjoints}{n_k}
\newcommand{\joints}{\textbf{J}}
\newcommand{\jointregressor}{\mathcal{J}}

\newcommand{\posecoeff}{\boldsymbol{\theta}}
\newcommand{\posedim}{{3\numjoints+3}}
\newcommand{\posespace}{\mathcal{P}}
\newcommand{\posespaceexpl}{\mathbb{R}^{\posedim}}

\newcommand{\expcoeff}{\boldsymbol{\psi}}
\newcommand{\expdim}{{\left| \expcoeff \right|}}
\newcommand{\expspace}{\mathcal{E}}
\newcommand{\expspaceexpl}{\mathbb{R}^{\expdim}}

\newcommand{\numverts}{n_v}
\newcommand{\numfaces}{n_t}
\newcommand{\template}{\mathbf{T}}

\newcommand{\blendweight}{w}
\newcommand{\blendweights}{\mathcal{W}}
\newcommand{\blendweightsdim}{\left| \numjoints \times \numverts \right|}
\newcommand{\lbs}{\text{LBS}}

\newcommand{\verts}{\mathbf{V}}
\newcommand{\faces}{\mathbf{F}}
\newcommand{\vsmplx}{M}
\newcommand{\matsmplx}{\mathbf{M}}
\newcommand{\offsets}{\mathbf{O}}
\newcommand{\offset}{\vect{o}}

\newcommand{\offsetmodel}{{F_{d}}}
\newcommand{\texmodel}{{F_{t}}}
\newcommand{\clothmodel}{{F_{c}}}
\newcommand{\defmodel}{{F_{m}}}

\newcommand{\image}{I}
\newcommand{\imagemask}{S} %
\newcommand{\numframes}{n_f}
\newcommand{\framenum}{f}
\newcommand{\cam}{\vect{p}}
\newcommand{\meshrender}{\mathcal{R}_m}
\newcommand{\lossweight}[1]{\lambda_{#1}}

\newcommand{\volrender}{\mathcal{R}_v}
\newcommand{\col}{\vect{c}}
\newcommand{\rayrender}{C}
\newcommand{\density}{\sigma}
\newcommand{\ray}{R}
\newcommand{\numsamples}{n_s}

\newcommand{\landmark}{\textbf{k}}

\newcommand{\albedo}{A}
\newcommand{\albedocoeffs}{\boldsymbol{\alpha}}
\newcommand{\albedodim}{\left| \albedocoeffs \right|}
\newcommand{\normalcoeffs}{\boldsymbol{\nu}}
\newcommand{\normaldim}{\left| \normalcoeffs \right|}

\newcommand{\uvsize}{d}

\newcommand{\openpose}{\mbox{OpenPose}\xspace}
\newcommand{\cuda}{\mbox{CUDA}\xspace}
\newcommand{\pytorch}{\mbox{PyTorch}\xspace}
\newcommand{\mocap}{\mbox{MoCap}\xspace}
\newcommand{\inthewild}{\mbox{in-the-wild}\xspace}
\newcommand{\twoD}{2D\xspace}
\newcommand{\threeD}{3D\xspace}
\newcommand{\sixD}{6D\xspace}

\begin{abstract}
While recent work has shown progress on extracting clothed 3D human avatars from a single image, video, or a set of 3D scans, several limitations remain.
Most methods use a holistic representation to jointly model the body and clothing, which means that the clothing and body cannot be separated for applications like virtual try-on.
Other methods separately model the body and clothing, but they require training from a large set of 3D clothed human meshes obtained from 3D/4D scanners or physics simulations.
Our insight is that the body and clothing have different modeling requirements. 
While the body is well represented by a mesh-based parametric 3D model, implicit representations and neural radiance fields are better suited to capturing the large variety in shape and appearance present in clothing. 
Building on this insight, we propose \modelname (\modelnamelong), a hybrid model combining a mesh-based body with a neural radiance field. 
Integrating the mesh into the volumetric rendering in combination with a differentiable rasterizer enables us to optimize \modelname directly from monocular videos, without any 3D supervision. 
The hybrid modeling enables \modelname to 
(i) animate the clothed body avatar by changing body poses (including hand articulation and facial expressions),
(ii) synthesize novel views of the avatar, and
(iii) transfer clothing between avatars in virtual try-on applications.
We demonstrate that \modelname reconstructs clothing with higher visual quality than existing methods, that the clothing deforms with changing body pose and body shape, and that clothing can be successfully transferred between avatars of different subjects.
The code and models are available at \url{https://github.com/YadiraF/SCARF}.

\end{abstract}

\begin{teaserfigure}
\centering
\includegraphics[width=7.2in]{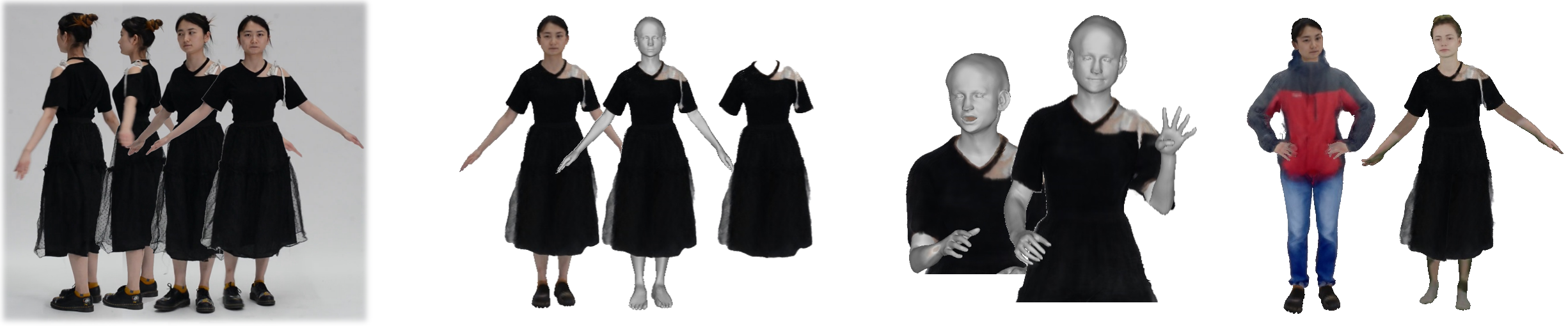}
\small{\hspace{1.2cm} (a) Input monocular video \hspace{1.2cm} (b) Disentangled captures from \modelname \hspace{0.8cm} (c) Animatable face and hands \hspace{0.8cm} (d) Clothing transfer \hspace{0.4cm}}
\caption{Given a monocular video (a), our method (\modelname) builds an avatar where the body and clothing are disentangled (b). The body is represented by a traditional mesh, while the clothing is captured by an implicit neural representation. \modelname enables animation with detailed control over the face and hands (c) as well as clothing transfer between subjects (d).}
\label{fig:teaser}
\end{teaserfigure}


\maketitle

\section{Introduction}
Realistic avatar creation is one of the key enablers of the metaverse, and it supports many applications in virtual presence, fitness, digital fashion, and entertainment. 
Traditional ways to build avatars require either complex capture systems or manual design by artists, both of which are time-consuming and inefficient for large-scale avatar creation. To address this, previous work explores more practical ways to create avatars directly from single RGB images or monocular videos, which are more accessible to consumers. 

The majority of work (e.g., \cite{Kanazawa2018_hmr, Kolotouros2019_SPIN, Pavlakos2019_smplifyx, Rong2021Frankmocap, Choutas2020ExPose, Zanfir2021_HUND, PIXIE:3DV:2021}) creates 3D human body avatars from images by estimating parameters of statistical 3D mesh models such as SCAPE~\cite{anguelov2005scape}, Adam~\cite{Joo2018Adam}, SMPL/SMPL-X \cite{SMPL:2015, Pavlakos2019_smplifyx}, GHUM~\cite{xu2020ghum}, or STAR~\cite{osman2020star}, or implicit surface models like imGHUM~\cite{Alldieck2021_imGHUM} and LEAP \cite{LEAP:CVPR:2021}. 
As these models are trained from minimally clothed body scans, they are unable to capture clothing shape and appearance variations, which require a more flexible representation.

Methods that recover clothed bodies from images are instead 
trained with a large set of 3D clothed human scans \cite{saito2019pifu, saito2020pifuhd, xiu2022icon}, or optimize the clothed avatar directly from multi-view images or videos \cite{jiang2022selfrecon, peng2022animatable, xu2021h, chen2021animatable, peng2021neural, liu2021neural, peng2021animatable}. 
To handle the complex topology of different clothing types, these methods model the body and clothing with a holistic implicit representation. 
Hence, hands and faces are typically poorly reconstructed and are not articulated.
Additionally, holistic models of the body and clothing do not permit virtual try-on applications, which require the body and clothing to be represented separately. 
While \ac{NeRF} is able to model the head well (e.g., \cite{Hong2022headnerf}), it remains unclear how to effectively combine such a part-based model with a clothed body representation.

Some methods treat the body and clothing separately with a layered representation, where clothing is modeled as a layer on top of the body \cite{jiang2020bcnet, zhu2020deep, corona2021smplicit, xiang2021modeling}.
These methods require large datasets of 3D clothing scans for training, but still lack generalization to diverse clothing types. Furthermore, given an RGB image, they recover only the geometry of the clothed body without appearance information \cite{jiang2020bcnet, zhu2020deep, corona2021smplicit}. 
Similarly, \citet{xiang2021modeling} require multi-view video data and accurately registered 3D clothing meshes to build a subject-specific avatar; their method is not applicable to loose clothing  like skirts or dresses.

Our goal is to go beyond existing work to capture realistic avatars from monocular videos that have detailed and animatable hands and faces as well as clothing that can be easily transferred between avatars.
We observe that the body and clothing have different modeling requirements. 
Human bodies have similar shapes that can be modeled well by a statistical mesh model. 
In contrast, clothing shape and appearance are much more varied, thus require more flexible 3D representations that could handle changing topologies and transparent materials.  
With these observations, we propose \modelname (\modelnamelong), a hybrid representation combining a mesh with a \ac{NeRF}, to capture disentangled clothed human avatars from monocular videos. 
Specifically, we use \smplx to represent the human body and a \ac{NeRF} on top of the body mesh to capture clothing of varied topology.
There are four main challenges in building such a model from monocular video. 
First, \modelname must accurately capture human motion in monocular video and relate the body motion to the clothing.
The \ac{NeRF} is modeled in canonical space, and we use the skinning transformation from the \smplx body model to deform points in observation space to the canonical space. %
This requires accurate estimates of body shape and pose for every video frame.
We estimate body pose and shape parameters with PIXIE \cite{PIXIE:3DV:2021}. 
However, these estimates are not accurate enough, resulting in blurry reconstructions. 
Thus, we refine the body pose and shape during optimization. 
Second, the cloth deformations are not fully explained by the \smplx skinning, particularly in the presence of loose clothing.
To overcome this, we learn a non-rigid deformation field to correct clothing deviations from the body. 
Third, \modelname's hybrid representation, combining a \ac{NeRF} and a mesh, requires customized volumetric rendering. 
Specifically, rendering the clothed body must account for the occlusions between the body mesh and the clothing layer. 
To integrate a mesh into volume rendering, we sample a ray from the camera's optical center until it intersects the body mesh, and accumulate the colors along the ray up to the intersection point with the colored mesh surface. 
Fourth, to disentangle the body and clothing, we must prevent the \ac{NeRF} from capturing all image information including the body. 
To that end, we use clothing segmentation masks to penalize the \ac{NeRF} outside of clothed regions. 

In summary, \modelname automatically creates a 3D clothed human avatar from monocular video (Fig.~\ref{fig:teaser}) with disentangled clothing on top of the human body.
\modelname offers the best of two worlds by combining different representations -- a 3D parametric model for the body and a \ac{NeRF} for the clothing.
Based on \smplx, the reconstructed avatar offers animator control over body shape, pose, hand articulation, and facial expression. 
Since \modelname factors clothing from the body, the clothing can be extracted and transferred between avatars, enabling applications such as virtual try-on.

\section{Related work}
\begin{figure*}[t]
	\centering
	\includegraphics[width=\textwidth, trim={0cm, 0cm, 0cm, 0cm}]{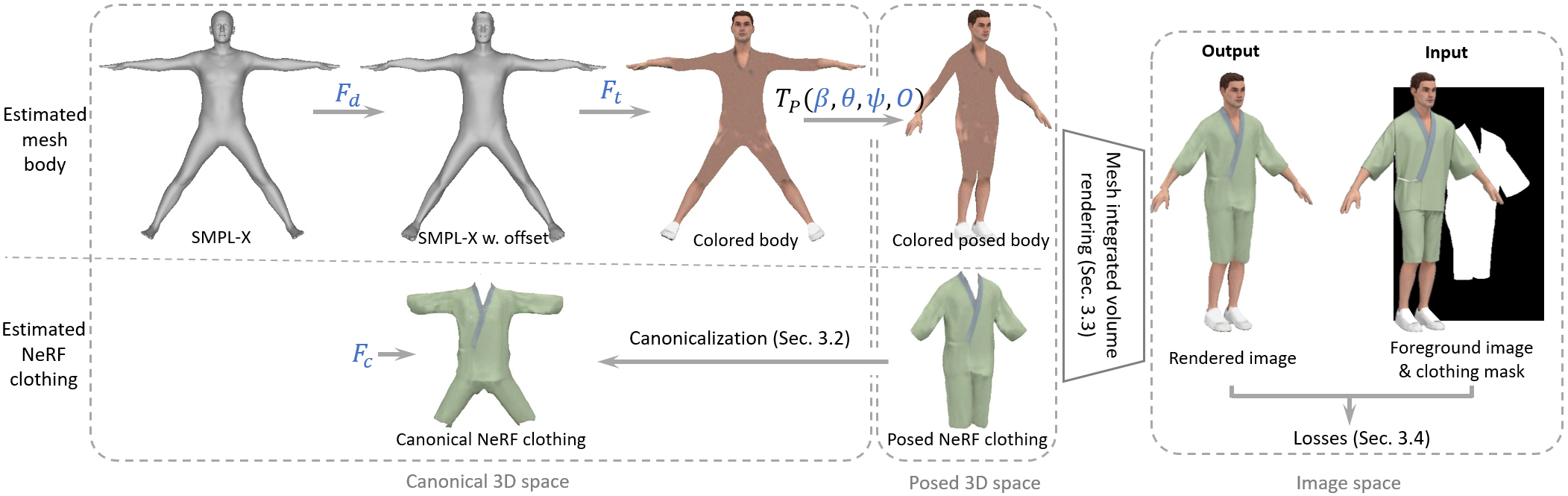}  
	\caption{\modelname takes monocular RGB video and clothing segmentation masks as input, and outputs a human avatar with separate body and clothing layers. Blue letters indicate optimizable modules or parameters.}
	\label{fig:pipeline}
\end{figure*}

\qheading{3D Bodies from images} 
The 3D surface of a human body is typically represented by a learned statistical 3D model \cite{anguelov2005scape, Joo2018Adam, SMPL:2015, osman2020star, Pavlakos2019_smplifyx, xu2020ghum, Alldieck2021_imGHUM}. 
Numerous optimization and regression methods have been proposed to compute 3D shape and pose parameters from images, videos, and scans.
See \cite{tian2022hmrsurvey, Liu2021hmrsurvey} for recent surveys.
We focus on methods that capture full-body pose and shape, including the hands and facial expressions \cite{Pavlakos2019_smplifyx, Choutas2020ExPose, PIXIE:3DV:2021, Xiang2019, Rong2021Frankmocap, Zhou2021, xu2020ghum}.
Such methods, however, do not capture hair, clothing, or anything that deviates the body. 
Also, they rarely recover texture information, due to the large geometric discrepancy between the clothed human in the image and captured minimal clothed body mesh. 
Unlike these prior works, we consider clothing as an important component and capture both the parametric body and non-parametric clothing from monocular videos.

\qheading{Capturing clothed humans from images} 
Clothing is more complex than the body in terms of geometry, non-rigid deformation, and appearance, making the capture of clothing from images challenging. 
Mesh-based methods to capture clothing often use additional vertex offsets relative to the body mesh \cite{alldieck2018video, alldieck20183DV, lazova3dv2019, alldieck2019learning, alldieck2019tex2shape,  ma2020learning, CAPE:CVPR:20, jin2020pixel}. 
While such an approach works well for clothing that is similar to the body, it does not capture clothing of varied topology like skirts and dresses.
To handle clothing shape variations, recent methods exploit non-parametric models. For example, \cite{huang2020arch, he2021arch++, saito2019pifu, saito2020pifuhd, xiu2022icon, zheng2021pamir} extract pixel-aligned spatial features from images and map them to an implicit shape representation.
To animate the captured non-parametric clothed humans, \citet{yang2021s3} predict skeleton and skinning weights from images to drive the representation.
 Although such non-parametric models can capture various clothing styles much better than mesh-based approaches, faces and hands are usually poorly recovered due to the lack of a strong prior on how the human body should be. In addition, such approaches typically require a large set of manually cleaned 3D scans as training data. 
Recently, various methods recover 3D clothed humans directly from multi-view or monocular RGB videos \cite{su2021nerf, weng2022humannerf, liu2021neural, peng2021neural, chen2021animatable, peng2021animatable, jiang2022selfrecon, peng2022animatable}. They optimize avatars from image information using implicit shape rendering \cite{liu2020dist, yariv2020multiview, yariv2021volume, niemeyer2020differentiable} or volume rendering~\cite{mildenhall2020nerf}, no 3D scans are needed. 
Although these approaches demonstrate impressive performance, hand gestures and facial expressions are difficult to capture and animate due to the lack of model expressivity and controllability. 
Unlike previous work, we capture clothing as a separate component on top of the body. With such a formulation, we use models tailored specifically to bodies and clothing, enabling applications such as virtual try-on and clothing transfer.

\qheading{Capturing both clothing and body} 
Several methods model clothing as a separate layer on top of the human body. 
They use training data produced by physics-based simulations~\cite{bertiche2020cloth3d, santesteban2019learning, vidaurre2020fully, patel2020tailornet} or require template meshes fit to  3D scans \cite{pons2017clothcap, xiang2021modeling, tiwari2020sizer, chen2021tightcap}. 
It is a much harder problem to recover the body and clothing from images alone, where 3D data is not provided. 
\citet{jiang2020bcnet} and \citet{zhu2020deep} train a multi-clothing model on 3D datasets with various clothing styles. Then during inference, a trained network produces the 3D clothing as a separate layer by recognizing and predicting the clothing style from an image. \citet{zhu2022registering} fit template meshes to non-parametric 3D reconstructions. 
While these methods recover the clothing and body from images, they are limited in visual fidelity, as they do not capture clothing appearance.
Additionally, methods with such predefined clothing style templates can not easily handle the real clothing variations, limiting their applications. 
In contrast, \citet{corona2021smplicit} represent clothing layers with deep unsigned distance functions \cite{chibane2020ndf}, and learn the clothing style and clothing cut space with an auto-decoder. 
Once trained, the clothing latent code can be optimized to match image observations, but it produces over-smooth results without detailed wrinkles. 
Instead, \modelname models the clothing layer with a neural radiance field, and optimizes the body and clothing layer from scratch instead of the latent space of a learned model. 
Therefore, \modelname produces avatars with higher visual fidelity (see Section~\ref{sec:experiments}).

\section{Method}

\modelname extracts a clothed 3D avatar from a monocular video.
\modelname enables us to synthesize novel views of the reconstructed avatar, and to animate the avatar with \smplx identity shape and pose control. 
The disentanglement of body and clothing further enables us to transfer clothing between subjects for virtual try-on applications. 

\qheading{Key idea}
\modelname is grounded in the observation that statistical mesh models can represent human bodies well, but are ill-suited for clothing due to the large variation in clothing shape and topology (e.g., open \& closed jackets, shirt, trousers, and skirts cannot be modeled with meshes of the same topology). 
Instead, \ac{NeRF} \cite{mildenhall2020nerf} offers more flexibility for modeling clothing, but is less appropriate for bodies where good models already exist.
In particular, body \ac{NeRF}s often lack facial details, poorly reconstruct hands, and
lack fine-grained control of hand articulation and facial expression \cite{peng2022animatable, su2021nerf, peng2021neural, chen2021animatable}.
Motivated by the strengths and weaknesses of the different representations, we use a hybrid representation that combines the strengths of body mesh models (specifically \smplx) with the flexibility of \ac{NeRF}s; see Figure~\ref{fig:pipeline} for an overview.

\subsection{Hybrid Representation}
We define the clothed body model in a canonical space, where body and clothing are represented separately.

\qheading{Body representation} 
We represent the body with the expressive body model, \smplx \cite{Pavlakos2019_smplifyx}, which captures whole-body shape and pose variations, including finger articulation, and facial expressions. 
Given parameters for identity body shape $\shapecoeff \in \shapespaceexpl$, pose $\posecoeff \in \posespaceexpl$, and facial expression $\expcoeff \in \expspaceexpl$, \smplx is defined as a differentiable function $\vsmplx(\shapecoeff, \posecoeff, \expcoeff) \rightarrow (\verts, \faces)$ that outputs a \threeD human body mesh with $\numverts$ vertices $\verts \in \mathbb{R}^{\numverts \times 3}$, and $\numfaces$ faces  $\mathbf{F} \in \mathbb{R}^{\numfaces \times 3}$. 
To increase the flexibility of the model, we add an additional set of vertex offsets $\offsets \in \mathbb{R}^{\numverts \times 3}$ to capture localized geometric details, and define the model as
\begin{equation}
    \vsmplx(\shapecoeff, \posecoeff, \expcoeff, \offsets) = \lbs(T_P(\shapecoeff, \posecoeff, \expcoeff, \offsets), \joints(\shapecoeff), \posecoeff, \blendweights),
    \label{eq:smplx}
\end{equation}
with $\numjoints$ shape dependent joints $\joints \in \mathbb{R}^{\numjoints \times 3}$, which are a function of body shape.
The linear blend skinning function $\lbs$ uses blend skinning weights $\blendweights \in \mathbb{R}^{\numjoints \times \numverts}$, and 
\begin{equation}
    T_P(\shapecoeff, \posecoeff, \expcoeff, \offsets) = \template + \offsets + B(\shapecoeff, \posecoeff, \expcoeff),
\end{equation}
where $\template \in \mathbb{R}^{\numverts \times 3}$ is a template in rest pose, and the blend shapes
\begin{equation}
    B(\shapecoeff, \posecoeff, \expcoeff) = B_{S}(\shapecoeff;\shapespace) + B_{P}(\posecoeff;\posespace) + B_{E}(\expcoeff;\expspace).
\end{equation}
Here, $B_{S}(\shapecoeff;\shapespace): \shapespaceexpl \rightarrow \mathbb{R}^{\numverts \times 3}$ are the identity blend shapes, $B_{P}(\posecoeff;\posespace): \posespaceexpl \rightarrow \mathbb{R}^{\numverts \times 3}$ are the pose blend shapes, and $B_{E}(\expcoeff;\expspace): \expspaceexpl \rightarrow \mathbb{R}^{\numverts \times 3}$ are the expression blend shapes with the learned identity $\shapespace$, pose $\posespace$, and expression $\expspace$ subspaces. 

Specifically, given a template vertex $\vect{t}_i$, the vertex $\vect{v}_i$ ($\vect{t}_i$ and $\vect{v}_i$ are column vectors in homogeneous coordinates) is computed as $\vect{v}_i = \matsmplx_i(\shapecoeff, \posecoeff, \expcoeff, \offsets) \vect{t}_i$, with $\matsmplx_i(.) \in \mathbb{R}^{4 \times 4}$ as 
\begin{equation}
    \matsmplx_i(\shapecoeff, \posecoeff, \expcoeff, \offsets) = \left(\sum_{k=1}^{\numjoints} \blendweight_{k,i} G_k(\posecoeff, \joints)\right) 
    \begin{bmatrix}
        \mathbf{I}  & \offset_i + B_i(\shapecoeff, \posecoeff, \expcoeff) \\
        \vect{0}^T & 1
    \end{bmatrix},
    \label{eq:transformation}
\end{equation}
where $\blendweight_{k,i}$ is a blend weight element of $\blendweights$, $G_k(\posecoeff, \joints) \in \mathbb{R}^{4 \times 4}$ is the world transformation of joint $k$, $\mathbf{I} \in \mathbb{R}^{3 \times 3}$ is the identity matrix, and $\offset_i$ and $B_i(.)$ are the elements of the i-the vertex of $\offsets$ and $B(.)$, respectively. 
For more details regarding the \smplx formulation, we refer to \citet{Pavlakos2019_smplifyx}.

To capture more geometric details, we use an upsampled version of \smplx with $\numverts=38,703$ vertices and $\numfaces=77,336$ faces. 
We obtain this by subdividing a quad version of the model's template, and upsampling the blend shape bases and skinning weights using barycentric coordinates obtained from the upsampled template. 
As the upsampling does not increase the variability of the model, we add additional learnable vertex offsets $\offsets$ for each subject. 
Similar to \citet{Grassal2022}, we use implicit models $\offsetmodel: \vect{t} \rightarrow \offset$ to describe the offset from every vertex $\vect{t}$ of $\template$, and $\texmodel: \vect{t} \rightarrow \col$ to predict the RGB color of every vertex $\vect{t}$.

\qheading{Clothing representation}
Due to the large variety of clothing in in-the-wild videos, we represent clothing using \ac{NeRF} \cite{mildenhall2020nerf} due to its ability to handle diverse topologies and transparent cloth materials. 
Following previous work (e.g., \cite{chen2021animatable,peng2021animatable}), we define the \ac{NeRF} model in canonical space as $\clothmodel: \vect{x}^{c} \rightarrow (\col, \density)$ to predict RGB color $\col$ and density $\density$ for each query point $\vect{x}^{c} \in \mathbb{R}^3$. %
Note that unlike previous work that models entire clothed bodies with a \ac{NeRF} (e.g., \cite{peng2021neural,chen2021animatable,peng2021animatable, liu2021neural, weng2022humannerf}), we only represent clothing part with a \ac{NeRF}. The whole clothed body then consists of an implicit representation \ac{NeRF} for clothing and an explicit surface representation for the underlying human body. 

The skinning articulation of a body model like \smplx is not sufficient to model pose-dependent clothing deformations.
Following previous work (\cite{liu2021neural, peng2022animatable, weng2022humannerf}), to model pose-dependent effects, we learn a deformation function $\defmodel: \mathbb{R}^6 \rightarrow \mathbb{R}^3$ in the canonical space to model the residual non-rigid deformation. %
Specifically, given a body mesh $\vsmplx(\shapecoeff, \posecoeff, \expcoeff, \offsets) \rightarrow \verts$ and a point $\vect{x}$ and $\vect{x}^{c}$ in observation space and canonical space, respectively, we optimize the weights of an MLP $\defmodel: (\vect{x}^c, \vect{v}_{\text{nn}(x)}^{p}) \rightarrow \vect{d}^{c}$, where $\text{nn}(x)$ is the index of the nearest neighbor vertex of $\vect{x}$ in $\verts$.
This MLP conditions $\vect{x}^c$ on a vertex $\vect{v}^{p}$ from the posed mesh $\vsmplx(\vect{0}, \posecoeff, \vect{0}, \vect{0}) \rightarrow \verts^{p}$.
Instead of $\vect{x}^c$, the displaced point $\vect{x}^c + \vect{d}^{c}$ in canonical space is then input to $\clothmodel$.

\subsection{Canonicalization}
To model the body and clothing in canonical space, we need to transfer points in observation space to the canonical space.
Following \citet{chen2021animatable}, we use the inverse transformation of the underlying \smplx model to transform from the pose $\posecoeff$ in observation space to the ``star-like'' body pose $\posecoeff^{c}$ (Fig.~\ref{fig:pipeline}) in canonical space. 

As the transformation between canonical space and observation space (Eq.~\ref{eq:transformation}) is only defined for surface vertices of the body model, \citet{zheng2021pamir} and \citet{chen2021animatable} generalize the model transformation to the entire space. 
Formally, given a body mesh $\vsmplx(\shapecoeff, \posecoeff, \expcoeff, \offsets) \rightarrow \verts$ and a point  $\vect{x}$ (in homogeneous coordinates) in observation space, $\vect{x}$ is transferred to canonical space with
\begin{equation}
    \sum_{\vect{v}_i \in \mathcal{N}\left(\vect{x}\right)} \frac{\omega_i(\vect{x})}{\omega(\vect{x})} 
        \mathbf{M}_{i}(\vect{0}, \posecoeff^{c}, \vect{0}, \vect{0})
        \left(\mathbf{M}_{i}(\shapecoeff, \posecoeff, \expcoeff, \offsets)\right)^{-1}\textbf{x} \rightarrow \textbf{x}^{c},
\end{equation}
where $\mathcal{N}\left(\vect{x}\right)$ is the set of nearest neighbor vertices of $\vect{x}$ in $\verts$. 
Further, the transformations are weighted with
\begin{equation}
    \begin{aligned}
        \omega_i(\vect{x})  &=\exp \left(-\frac{\left\|\vect{x}-\vect{v}_{i}\right\|_2 \left\|\mathbf{\blendweight}_{\text{nn}(x)}-\mathbf{\blendweight}_{i}\right\|_2}{2 \sigma^{2}}\right), \text{ and}  \\
        \omega(\vect{x}) &=\sum_{\vect{v}_i \in \mathcal{N}(\vect{x})} \omega_i(\vect{x}),   
    \end{aligned}
\end{equation}
where $\text{nn}(x)$ is the index of the nearest neighbor vertex of $\vect{x}$ in $\verts$, $\mathbf{\blendweight}_i \in \mathbb{R}^{\numjoints}$ are the blend weights of $\vect{v}_{i}$, and $\sigma$ is a constant weight.

\subsection{Mesh Integrated Volume Rendering}
\label{subsec:mesh_integrated_rendering}
\qheading{Camera} 
To reconstruct \smplx from images, we use a scaled-orthographic camera model $\cam = [s, \vect{t}^T]^T$ with isotropic scale $s \in \mathbb{R}$ and translation $\vect{t} \in \mathbb{R}^2$.

\qheading{Mesh rendering} 
Given geometry parameters ($\shapecoeff, \posecoeff, \expcoeff$), vertex offsets $\offsets$, colors $\texmodel: \vect{t}_i \rightarrow \col_i$ for every vertex in the upsampled \smplx  template, and camera information $\cam$, we render the colored mesh into an image as $\meshrender(\vsmplx(\shapecoeff, \posecoeff, \expcoeff, \offsets), \col, \cam)$, where $\meshrender$ denotes the differentiable rasterizer function.

\begin{figure}[t]
	\centering
	\includegraphics[width=\linewidth]{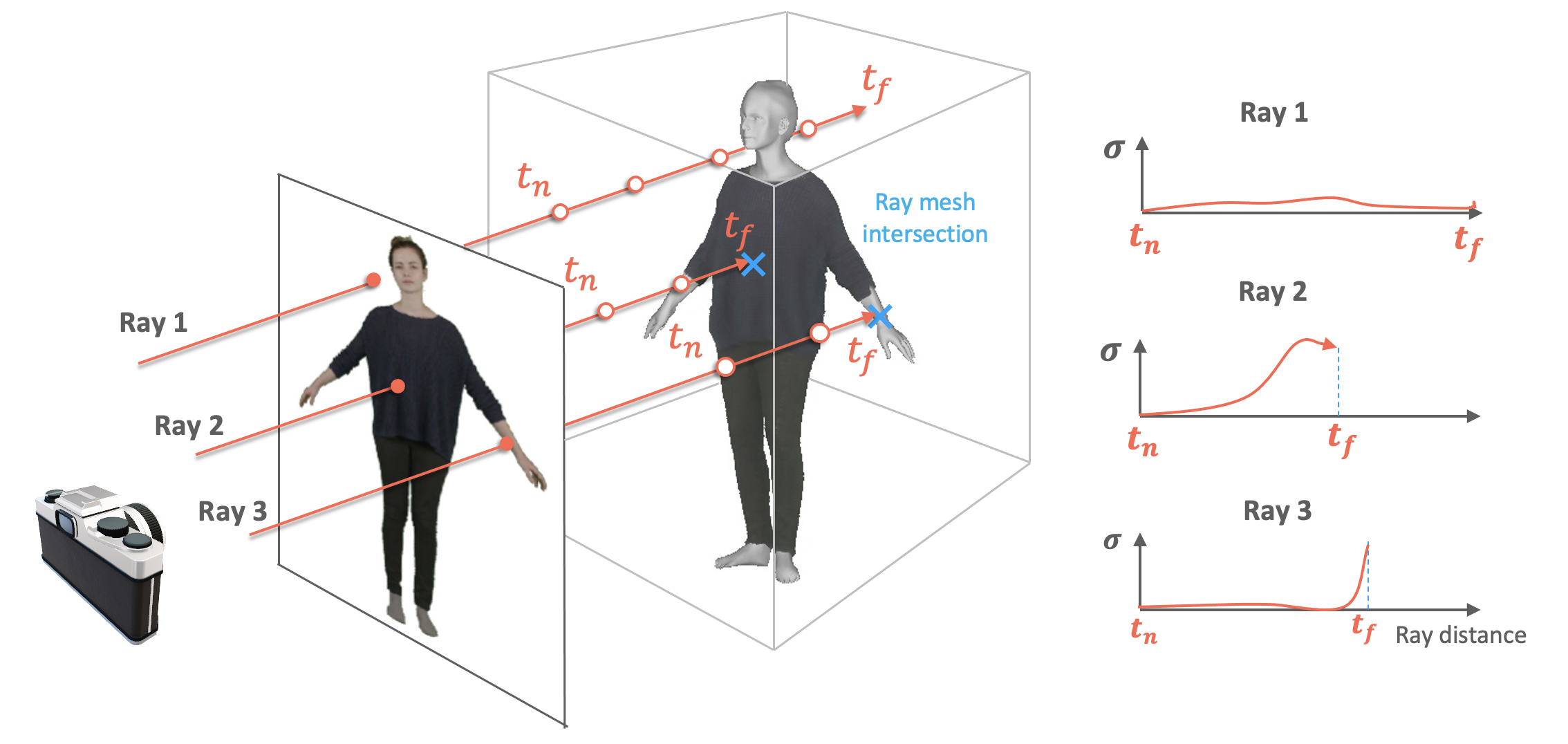} \\  
	\caption{Graphic illustration for mesh integrated volume rendering in Sec.~\ref{subsec:mesh_integrated_rendering}.}
	\label{fig:rendering}
\end{figure}

\qheading{Volume rendering} 
We follow \citet{mildenhall2020nerf} to use volumetric rendering. 
Given a camera ray $\ray(t) = \vect{o} + t \vect{d}$ with center $\vect{o} \in \mathbb{R}^3$ and direction $\vect{d} \in \mathbb{R}^3$, the rendering interval $t \in [t_n, t_f] \subset \mathbb{R}$ (near and far bounds) is evently split into $\numsamples$  bins.
A random sample $t_i$ ($1 \leq i \leq \numsamples$) from every bin is taken and the colors are aggregated across the ray samples $\ray(t_i) \rightarrow \vect{r}_i$. 
Unlike previous work, we integrate the body model, 
$\vsmplx(\shapecoeff,\posecoeff,\expcoeff,\offsets)$, into the volumetric rendering. 
Specifically, if $\ray(t)$ intersects $\vsmplx$, we set the $t_f$ such that $\ray(t_{\numsamples})$ is the intersection point with $\vsmplx$.
In this case, we use the mesh color instead of the \ac{NeRF} color $\col_{\numsamples}$ (see Fig.~\ref{fig:rendering}). 
Formally, the aggregated color is
\begin{equation}
    \begin{aligned}
        \rayrender(\mathbf{R}) &= \sum_{i=1}^{\numsamples-1} \alpha_i \col_{i} + \tau \col, \text{  with  } 
        \alpha_i = \gamma_i(1-\exp(-\density_{i}\delta_{i})), \text{  where } \\
        \gamma_i &= \prod_{j=1}^{i-1} \exp(-\density_{j}\delta_{j}), \text{  and  } \tau =  1-\sum_{i=1}^{\numsamples-1} \alpha_i,  \text{ and } \\
        \vect{c} &= 
        \begin{cases}
            \texmodel(\vect{r}_{\numsamples}^c), & \text{if } $\ray(t)$ \text{ intersects } $\vsmplx$  \\
            \col_{\numsamples},& \text{otherwise.} 
        \end{cases}        
    \end{aligned}
\end{equation} 
Here, $\delta_i = t_{i+1} - t_i$ is the distance between adjacent samples, $\mathbf{R} = \{ \vect{r}_1, \cdots, \vect{r}_{\numsamples}  \}$, $\clothmodel(\vect{r}_i) \rightarrow (\col_{i}, \density_{i})$, and $\vect{r}_{\numsamples}^c$ is the canonicalized $\vect{r}_{\numsamples}$.
For the scaled-orthographic camera, we use $\vect{o} = [o_x, o_y, 0]$ and $\vect{d} = [0,0,1]$ to compute the color of the pixel $[o_x, o_y]$.
We denote the image rendered by sampling rays for all image pixels as $\volrender$.

\subsection{Objectives}

Given a sequence of $\numframes$ images, $\image_f$ ($1\leq \framenum \leq \numframes)$, we  optimize $\shapecoeff$ and the weights of the MLPs $\offsetmodel, \clothmodel, \texmodel, \defmodel$ jointly across the entire sequence, and $\posecoeff_{\framenum} \text{ and } \cam_{\framenum}$ per frame.
The objective is
\begin{equation}
    L = L_{\text{recon}} + L_{\text{clothing}} + L_{\text{body}},
\end{equation}
with reconstruction loss $L_{\text{recon}}$, clothing segmentation loss $L_{\text{clothing}}$, and body loss $L_{\text{body}}$.
For simplicity, we omit the frame index $\framenum$ and the optimized parameters whenever possible. 
The sequence objective function is the sum over all frames.

\qheading{Reconstruction loss}
we minimize the difference between the rendered image and the input image as 
\begin{equation}
    L_{\text{recon}} = \lossweight{\text{vol}} L_{\delta}(\volrender-\image) + \lossweight{\text{mrf}} L_{\text{mrf}}(\volrender-\image),
\end{equation}
where $L_{\delta}$ is the Huber loss \cite{Huber1964}, and $L_{\text{mrf}}$ is an ID-MRF loss~\cite{Wang2018}.
While the Huber loss focuses on the overall reconstruction, the ID-MRF loss allows us to reconstruct more details as previously shown by \citet{Feng:SIGGRAPH:2021}.
Solely minimizing $L_{\text{recon}}$ results in a \ac{NeRF} that models the entire clothed body including the non-clothing regions.

\qheading{Cloth segmentation loss}
Our goal is to only capture clothing with $\clothmodel$ instead of modeling the entire clothed body. 
This requires us to disentangle body and clothing. 
Given a clothing mask $\imagemask_c$, which is $\vect{1}$ for every clothing pixel and $\vect{0}$ elsewhere, we minimize the clothing segmentation loss as
\begin{equation}
    L_{\text{clothing}} = \lossweight{\text{clothing}} \norm{\imagemask_v - \imagemask_c}_{1,1},
\end{equation}
with the rendered \ac{NeRF} mask $S_v$, which is obtained by sampling rays for all image pixels and computing per ray
\begin{equation}
    S(\mathbf{R}) = \sum_{i=1}^{\numsamples-1} \prod_{j=1}^{i-1} \exp(-\density_{j}\delta_{j})(1-\exp(-\density_{i}\delta_{i})). 
\end{equation}
Minimizing $L_{\text{clothing}}$ ensures that the aggregated density across rays (excluding the far bound) outside of clothing is $0$ and therefore nothing outside of the clothing mask is modeled by the \ac{NeRF}.

\qheading{Human body loss}
To further disentangle body and clothing, we must ensure that the body model does not capture clothing variations. 
For this purpose, we define different losses based on four observations.

First, the body mesh should match the masked image. 
Given a binary mask $\imagemask$ of the clothed body ($1$ for inside, $0$ elsewhere), we minimize the difference between the silhouette of the rendered body $\meshrender^{s}(\vsmplx, \cam)$ and the given mask as
\begin{equation}
    L_{\text{silhouette}} = \lossweight{\text{silhouette}} L_{\delta}(\meshrender^{s}(\vsmplx, \cam) - \imagemask).
\end{equation}

Second, the body mesh should match visible body parts.
Optimizing $L_{\text{silhouette}}$ only results in meshes that also fit the clothing, which is undesired especially for loose clothing (i.e., this leads to visible artifacts when transferring clothing between subjects). 
Instead, given a binary mask $\imagemask_b$ of the visible body parts ($1$ for body parts, $0$ elsewhere), we minimize a part-based silhouette loss
\begin{equation}
    L_{\text{bodymask}} = \lossweight{\text{bodymask}} L_{\delta}(\imagemask_b \odot \meshrender^{s}(\vsmplx, \cam) - \imagemask_b),
\end{equation}
and a part-based photometric loss
\begin{equation}
    L_{\text{skin}} = \lossweight{\text{skin}} L_{\delta}(\imagemask_b \odot (\meshrender(\vsmplx, \col, \cam)-\image)),
\end{equation}
to put special emphasis on fitting visible body parts. 

Third, the body mesh should stay within clothing regions, as
\begin{equation}
    L_{\text{inside}} = \lossweight{\text{inside}} L_{\delta}(ReLU(\meshrender^{s}(\vsmplx, \cam) - \imagemask_c)). 
\end{equation}

Fourth, the skin color of occluded body vertices should be similar to non-occluded regions. 
For this, we assume that hands are visible for some parts of the sequence, and minimize the difference between the body colors in occluded regions and the hand color as
\begin{equation}
    L_{\text{skininside}} = \lossweight{\text{skininside}} L_{\delta}(\imagemask_c \odot (\meshrender(\vsmplx, \col, \cam)-\vect{C}_{\text{hand}})),
\end{equation}
where $\vect{C} = [\col_{\text{hand}}^T, \cdots, \col_{\text{hand}}^T]^T \in \mathbb{R}^{\numverts \times 3}$ is the tiled average color $\col_{\text{hand}}$ of the hand vertices.

\qheading{Regularizion}
We regularize the reconstructed mesh surface as
\begin{equation*}
    L_{\text{reg}} = \lossweight{\text{edge}} L_{\text{edge}}(\vsmplx) 
                    + \lossweight{\text{offset}} \norm{\offsets}_{2,2}, 
\end{equation*}
where $L_{\text{edge}}$ is the relative edge loss \cite{Hirshberg2012_Coregistration} between the optimized body mesh w/ and w/o applied offsets. 
For the offset loss, we apply different weights on the body, hands and face region. For more details see the \supmat

Overall, the body loss is
\begin{equation}
    L_{\text{body}} = L_{\text{silhouette}} + L_{\text{bodymask}} + L_{\text{skin}} + L_{\text{skininside}} + L_{\text{inside}} + L_{\text{reg}}.
\end{equation}

\subsection{Implementation}
\modelname is implemented in PyTorch, with a built-in PyTorch3D rasterizer \cite{Ravi2020_PyTorch3D}, and optimized with Adam \cite{Kingma2015}. 
For each frame, we run PIXIE~\cite{PIXIE:3DV:2021} to initialize $(\shapecoeff, \posecoeff, \expcoeff)$, and $\cam$.
For datasets without provided silhouette masks, we compute $\imagemask$ with \cite{RobustVideoMatting}, and \cite{clothsegmentation} for $\imagemask_c$. 
Following \citet{mildenhall2020nerf}, we optimize both a coarse and a fine MLP to represent the \ac{NeRF}.
Our optimization pipeline has two stages. 
We first jointly optimize the canonical \ac{NeRF} to estimate the entire clothed body (i.e., without clothing segmentation) and refine the \smplx pose for $100k$ iterations with a learning rate of $5e-4$.
Then, we optimize the full model for another $50k$ iterations with learning rates of $1e-4$ for the \ac{NeRF} ($\clothmodel$ and $\defmodel$) and pose refinement, and $1e-5$ for the mesh color model ($\texmodel$) and the offset ($\offsetmodel$).
For more details about the implementation, please refer to the \supmat

\section{Experiments}
\label{sec:experiments}

\begin{figure}[t]
    \centering
    \includegraphics[width=0.98\columnwidth]{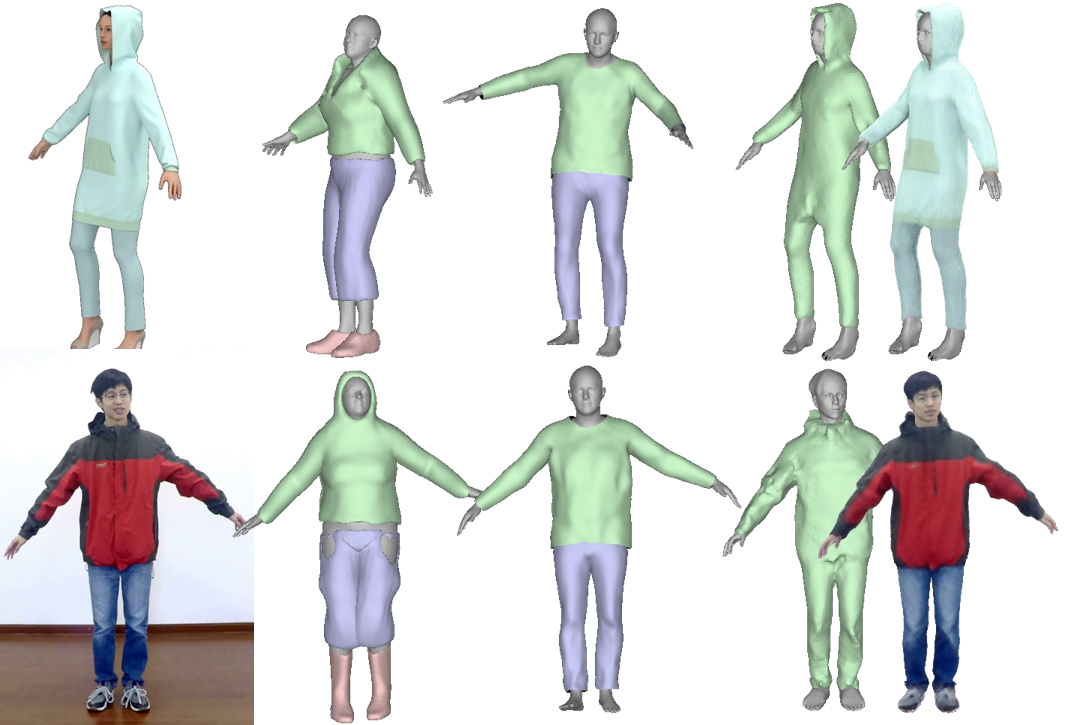}\\
    \small{Input image \hspace{1.5em} SMPLicit \hspace{3em} BCNet \hspace{5em} Ours \hspace{2em}}
    \vspace{-0.5em}
	\caption{Garment reconstruction comparison. \modelname reconstructs different clothing types more faithfully than SMPLicit \cite{corona2021smplicit} and BCNet \cite{jiang2020bcnet}.}
	\vspace{-1.0em}
	\label{fig:comparison_garment}
\end{figure}

\begin{figure}[t]
    \includegraphics[width=\linewidth]{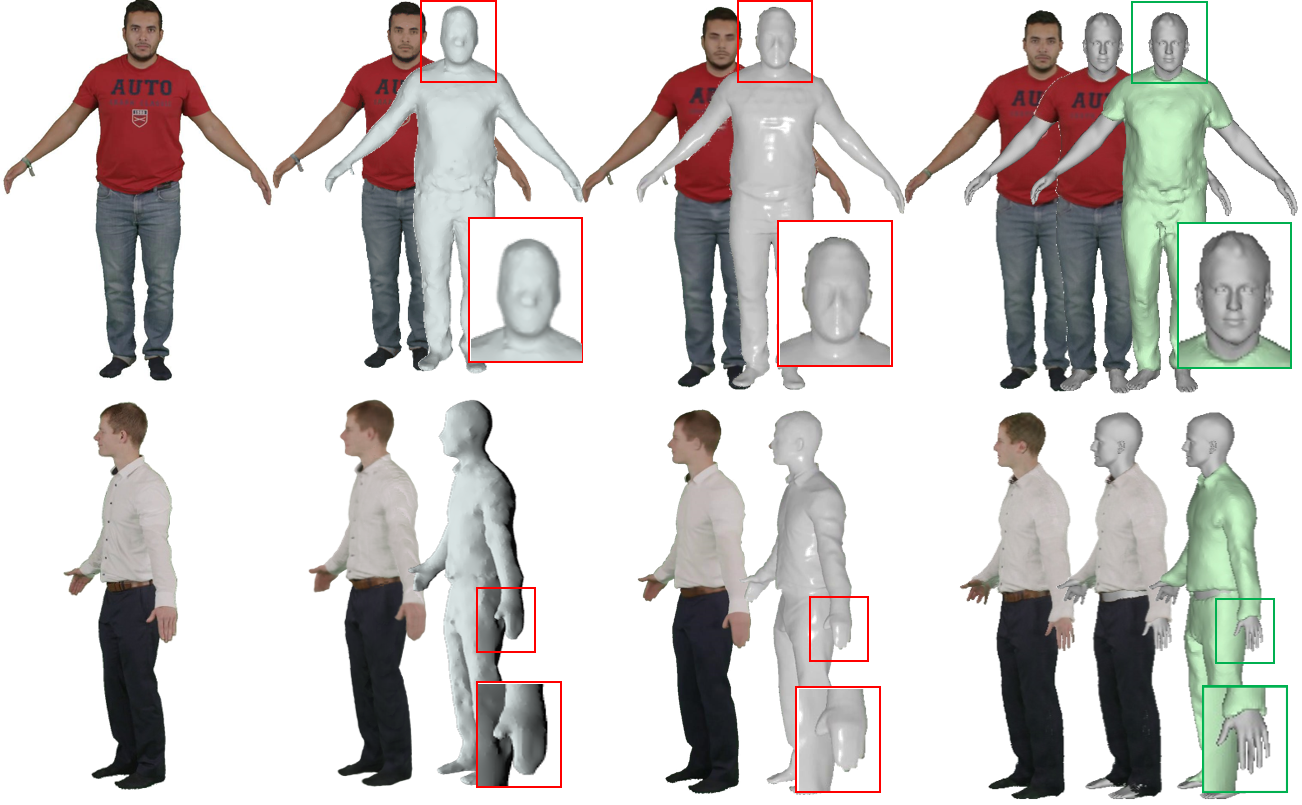}\\
    \raggedright
    \small{Reference image \hspace{1em} Anim-NeRF \hspace{2.5em} SelfRecon \hspace{4.5em} Ours }
    \vspace{-0.8em}
    \caption{Qualitative comparison with SelfRecon \cite{jiang2022selfrecon} and Anim-NeRF \cite{chen2021animatable} for reconstruction. While all methods capture the clothing with comparable quality, our approach has much more detailed face and hands due to the disentangled representation of clothing and body.}
	\label{fig:comparison_selfrecon}
\end{figure}

\setlength{\tabcolsep}{2.8pt}
\begin{table*}[ht]
	\begin{center}
	\resizebox{0.98\textwidth}{!}{
		\renewcommand{\arraystretch}{1.1}
		\begin{tabular}{c|ccccc|ccccc|ccccc}
            \toprule
			\multirow{2}{*}{Subject ID} & \multicolumn{5}{c|}  {PSNR$\uparrow$} & \multicolumn{5}{c|}{SSIM$\uparrow$} &
			\multicolumn{5}{c}{LIPIS$\downarrow$} \\
			\cline{2-16}
			                & NeRF  & SMPLpix   & NB    & Anim-NeRF    & Ours              & NeRF & SMPLpix & NB    & Anim-NeRF  & Ours           & NeRF & SMPLpix & NB    & Anim-NeRF  & Ours \\
			\midrule
			male-3-casual   & 20.64 & 23.74     & 24.94 & 29.37     & \textbf{30.59}    & .899 & .923  & .943 & .970   & \textbf{.977} & .101 & .022  & .033 & \textbf{.017} & .024 \\
			male-4-casual   & 20.29 & 22.43     & 24.71 & 28.37     & \textbf{28.99}    & .880 & .910  & .947 & .961   & \textbf{.970} & .145 & .031  & .042 & .027   & \textbf{.025} \\
			female-3-casual & 17.43 & 22.33     & 23.87 & 28.91     & \textbf{30.14}    & .861 & .929  & .950 & .974   & \textbf{.977} & .170 & .027  & .035 & \textbf{.022}   & .028 \\
			female-4-casual & 17.63 & 23.35     & 24.37 & 28.90     & \textbf{29.96}    & .858 & .926  & .945 & .968   & \textbf{.972} & .183 & .024  & .038 & \textbf{.017} & .026 \\
			\bottomrule
		\end{tabular}}
		\caption{Quantitative comparison of novel view synthesis on People-Snapshot \cite{alldieck2018video}.}
		\label{tab:comparison_snapshot}
	\end{center}
	\vspace{-0.2in}
\end{table*}

\subsection{Datasets}
We evaluate \modelname on sequences from People Snapshot \cite{alldieck2018video}, iPER \cite{lwb2019}, SelfRecon \cite{jiang2022selfrecon}, and self-captured data. 
For People Snapshot, we use the provided SMPL pose as initialization instead of running PIXIE \cite{PIXIE:3DV:2021}.
For each subject, we use around 100-150 images for optimization. 
See the \supmat for more details. 

\subsection{Comparisons} 
Our method can capture the body and clothing from image sequences, enabling novel view synthesis. 
Previous works either model whole clothed body from video or reconstruct cloth geometry from a single image after training with plentiful 3D scan data. 
So we compare our method with others on two tasks: novel view  synthesis and separate body and garment reconstruction from images. 

\qheading{Body and garment reconstruction} 
Similar to \modelname, SMPLicit \cite{corona2021smplicit} and BCNet \cite{jiang2020bcnet} separately model the body and clothing. 
Note that these methods and \modelname follow a different strategy. 
While they learn generative models from scans \cite{corona2021smplicit} or synthetic 3D data \cite{jiang2020bcnet} and then reconstruct the clothed body from a single image, \modelname extracts a clothed avatar from a video without 3D supervision. 
Figure~\ref{fig:comparison_garment} shows that \modelname reconstructs different clothing types more faithfully.

\qheading{Body and cloth modeling} 
We quantitatively compare to \ac{NeRF} \cite{omran2018neural}, SMPLpix \cite{SMPLpix:WACV:2020}, Neural Body \cite{peng2021neural} and Anim-NeRF \cite{chen2021animatable}, following the evaluation protocol of \cite{chen2021animatable}. 
Table \ref{tab:comparison_snapshot} shows that \modelname is more accurate than other methods under most metrics.
Figure~\ref{fig:comparison_selfrecon}  provides qualitative comparisons demonstrating that \modelname better reconstructs hand and face geometry compared to SelfRecon \cite{jiang2022selfrecon} and Anim-NeRF \cite{chen2021animatable}.

\begin{figure}[t]
    \includegraphics[width=\linewidth]{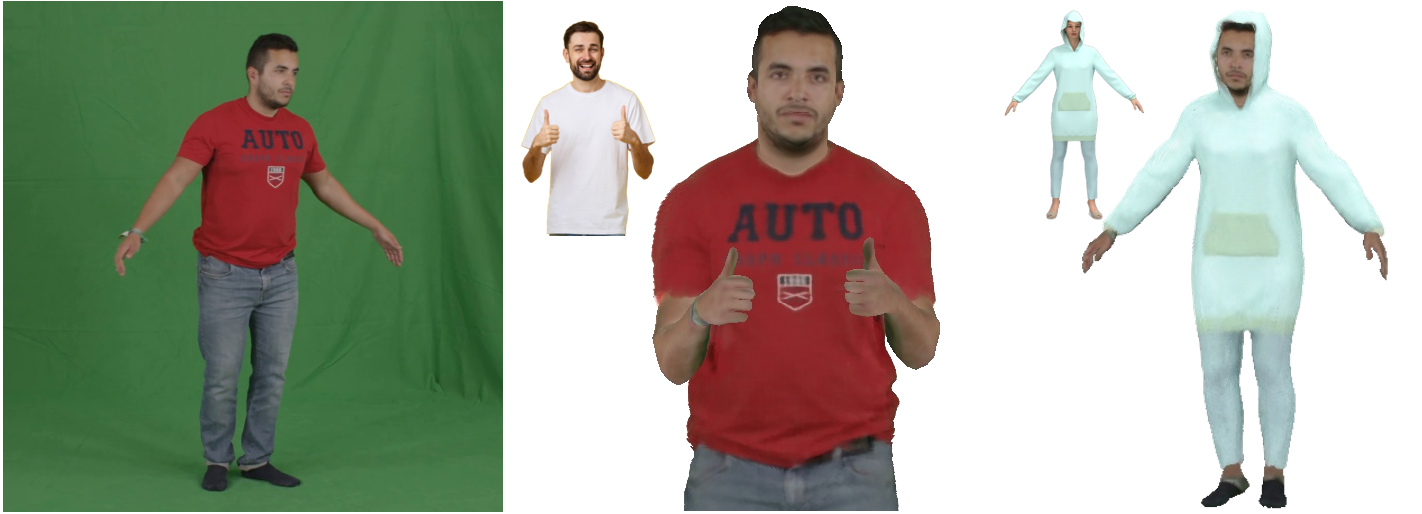}\\
    \small{Source subject \hspace{5em} Reposing \hspace{4em} Clothing transfer}
    \vspace{-0.8em}
    \caption{Applications of SCARF. The hybrid representation enables (middle) reposing with detailed control over the body pose and (right) dressing up the source subject with target clothing. The target pose and clothing are shown in the inset images.}
    \vspace{-0.5em}
    \label{fig:applications}
\end{figure}

\subsection{Applications}

\qheading{Animation}
Unlike previous methods that represent clothed bodies holistically, \modelname offers more fine grained control over body pose. 
Figure~\ref{fig:applications} shows reposing into novel poses.

\qheading{Cloth transfer} 
Figures \ref{fig:teaser} and \ref{fig:applications} and the \supmat show that our hybrid representation enables transfer of clothing between avatars.

\subsection{Ablation Experiments}
We run different ablation experiments to show the impact of different components of our hybrid representation (below), and to show the impact of the pose optimization in \supmat

\begin{figure}[t]
    \includegraphics[width=\columnwidth]{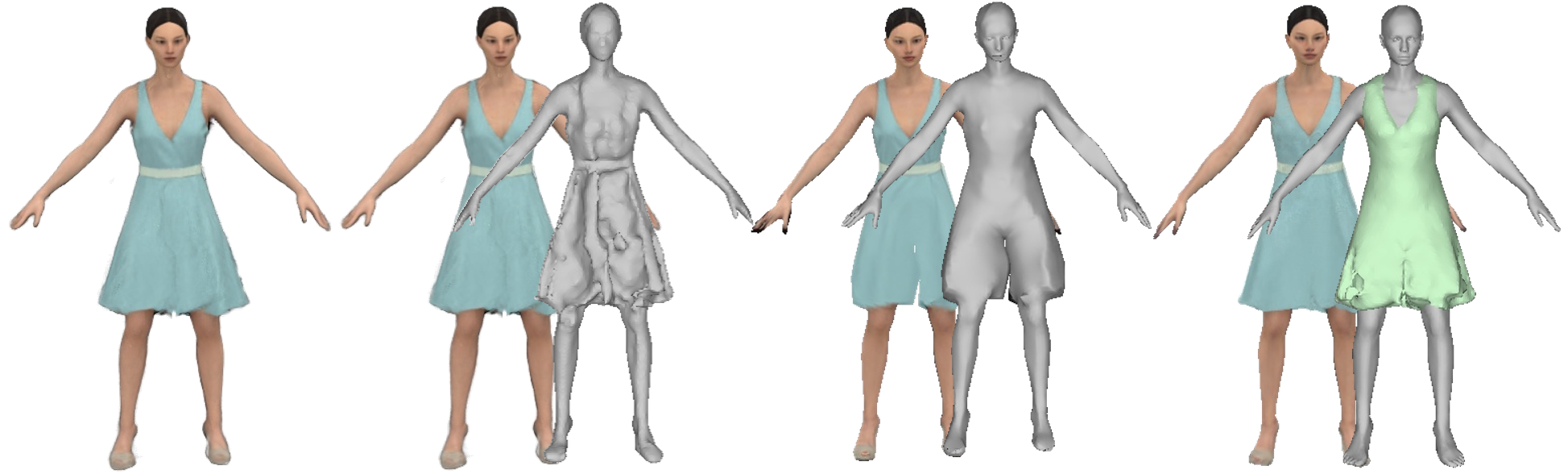}\\
    \raggedright
    \small{Reference image \hspace{2.5em} NeRF \hspace{3.3em} SMPL-X +D \hspace{4.2em} Ours}
     \vspace{-0.05in}
    \caption{Rendered images and extracted meshes from different components of \modelname.
    Our hybrid representation gives better estimated face, hand, and clothing geometry than vanilla \ac{NeRF} or a mesh-based representation. }
    \label{fig:ablation_representation}
 	\vspace{-1em}
\end{figure}

\qheading{Effect of representations}
\modelname consists of a \ac{NeRF} to represent clothing, and a mesh with vertex displacements. 
Figure~\ref{fig:ablation_representation} compares \ac{NeRF} to holistically represent body and clothing (i.e., \modelname w/o body-clothing segmentation) and mesh-only based representation (i.e., \modelname w/o \ac{NeRF}). 
Our hybrid representation is better able to estimate the face, hands, and complex clothing. 
Note that, unlike our hybrid representation, none of the existing body \ac{NeRF} methods is able to transfer clothing between avatars.

\section{Discussion and limitations}
\qheading{Segmentation}
\modelname requires body and cloth segmentation for training. 
Segmentation errors of the clothed body and background negatively impact the visual quality of the extracted avatar, and erroneous clothing segmentation results in poor separation of body and clothing. 
Enforcing temporal consistency by exploiting optical flow could improve the segmentation quality.

\qheading{Shoes \& hair}
Modeling hair, shoes, or other accessories with \ac{NeRF} would improve the visual quality of \modelname. 
We will explore alternative shoe and hair segmentation methods (e.g., \cite{Yang2020PRRCNN}) to extend \modelname.

\qheading{Geometric quality}
The strength of \ac{NeRF} is its visual quality and the ability to synthesize realistic images, even when the geometry is not perfect.
In contrast, recent SDF-based methods have demonstrated good geometric reconstruction (e.g., \cite{jiang2022selfrecon}). 
It may be possible to leverage their results to better represent the underlying clothed shape or to regularize \ac{NeRF}.

\qheading{Novel poses}
While \modelname generalizes to unseen poses, extreme poses result in artifacts (see \supmat).
Possible solutions include regularizing \ac{NeRF} during optimization or learning a generative model from many training examples of different people and poses.

\qheading{Pose initialization}
\modelname refines the body pose during optimization. However, it may fail if the initial pose is far from the right pose. 
Handling difficult poses where PIXIE~\cite{PIXIE:3DV:2021} fails requires a more robust 3D body pose estimator.

\qheading{Dynamics}
\modelname handles non-rigid cloth deformation with the pose-conditioned deformation model. 
While the global pose accounts for some deformation, the modeling of clothing dynamics as a function of body movement is the subject of future work.

\qheading{Lighting}
As with other \ac{NeRF} methods, we do not factor lighting and material properties.
This results in baked-in shading and the averaging of specular reflections across frames.
Factoring lighting from shape and material is a key next step to improve realism.

\qheading{Facial expressions}
\modelname uses the facial expressions estimated by PIXIE \cite{PIXIE:3DV:2021} which is unable to capture the full spectrum of emotions (cf.~\cite{EMOCA:CVPR:2021}).
Also, we have not fully exploited neural radiance fields to capture complex changes in facial appearance, e.g.~due to the mouth opening. We believe this is a promising future direction.

\section{Conclusion}
\modelname automatically extracts an animatable clothed 3D human avatar from a monocular video. 
Our key novelty is a hybrid representation that combines a mesh-based body model with a neural radiance field to separately model the body and clothing. 
This factored representation enables \modelname to transfer clothing between avatars, animate the body pose of the avatars including finger articulation, alter their body shape (see \supmat) and facial expression, and visualize them from unseen viewing directions. 
This property makes \modelname well suited to VR and virtual try-on applications.
Finally, \modelname outperforms existing avatar extraction methods from videos in terms of visual quality and generality.

\begin{acks}
We thank Sergey Prokudin, Weiyang Liu, Yuliang Xiu, Songyou Peng, Qianli Ma for fruitful discussions, and Peter Kulits, Zhen Liu, Yandong Wen, Hongwei Yi, Xu Chen, Soubhik Sanyal, Omri Ben-Dov, Shashank Tripathi for proofreading. 
We also thank Betty Mohler, 
Sarah Danes, Natalia Marciniak, Tsvetelina Alexiadis,
Claudia Gallatz, and Andres Camilo Mendoza Patino for their supports with data.
This work was partially supported by the Max Planck ETH Center for Learning Systems.

\qheading{Disclosure} MJB has received research
gift funds from Adobe, Intel, Nvidia, Meta/Facebook, and
Amazon. MJB has financial interests in Amazon, Datagen
Technologies, and Meshcapade GmbH. 
While TB is part-time employee of Amazon, this research was performed solely at, and funded solely by, MPI.
\end{acks}

\balance
\bibliographystyle{ACM-Reference-Format}
\bibliography{bibliography}


\begin{thebibliography}{75}


\ifx \showCODEN    \undefined \def \showCODEN     #1{\unskip}     \fi
\ifx \showDOI      \undefined \def \showDOI       #1{#1}\fi
\ifx \showISBNx    \undefined \def \showISBNx     #1{\unskip}     \fi
\ifx \showISBNxiii \undefined \def \showISBNxiii  #1{\unskip}     \fi
\ifx \showISSN     \undefined \def \showISSN      #1{\unskip}     \fi
\ifx \showLCCN     \undefined \def \showLCCN      #1{\unskip}     \fi
\ifx \shownote     \undefined \def \shownote      #1{#1}          \fi
\ifx \showarticletitle \undefined \def \showarticletitle #1{#1}   \fi
\ifx \showURL      \undefined \def \showURL       {\relax}        \fi
\providecommand\bibfield[2]{#2}
\providecommand\bibinfo[2]{#2}
\providecommand\natexlab[1]{#1}
\providecommand\showeprint[2][]{arXiv:#2}

\bibitem[Alldieck et~al\mbox{.}(2019a)]%
        {alldieck2019learning}
\bibfield{author}{\bibinfo{person}{Thiemo Alldieck}, \bibinfo{person}{Marcus
  Magnor}, \bibinfo{person}{Bharat~Lal Bhatnagar}, \bibinfo{person}{Christian
  Theobalt}, {and} \bibinfo{person}{Gerard Pons-Moll}.}
  \bibinfo{year}{2019}\natexlab{a}.
\newblock \showarticletitle{Learning to reconstruct people in clothing from a
  single RGB camera}. In \bibinfo{booktitle}{\emph{Proceedings of the IEEE/CVF
  Conference on Computer Vision and Pattern Recognition}}.
  \bibinfo{pages}{1175--1186}.
\newblock


\bibitem[Alldieck et~al\mbox{.}(2018a)]%
        {alldieck20183DV}
\bibfield{author}{\bibinfo{person}{Thiemo Alldieck}, \bibinfo{person}{Marcus
  Magnor}, \bibinfo{person}{Weipeng Xu}, \bibinfo{person}{Christian Theobalt},
  {and} \bibinfo{person}{Gerard Pons-Moll}.} \bibinfo{year}{2018}\natexlab{a}.
\newblock \showarticletitle{Detailed Human Avatars from Monocular Video}. In
  \bibinfo{booktitle}{\emph{International Conference on 3D Vision (3DV)}}.
\newblock


\bibitem[Alldieck et~al\mbox{.}(2018b)]%
        {alldieck2018video}
\bibfield{author}{\bibinfo{person}{Thiemo Alldieck}, \bibinfo{person}{Marcus
  Magnor}, \bibinfo{person}{Weipeng Xu}, \bibinfo{person}{Christian Theobalt},
  {and} \bibinfo{person}{Gerard Pons-Moll}.} \bibinfo{year}{2018}\natexlab{b}.
\newblock \showarticletitle{Video based reconstruction of 3d people models}. In
  \bibinfo{booktitle}{\emph{Proceedings of the IEEE Conference on Computer
  Vision and Pattern Recognition}}. \bibinfo{pages}{8387--8397}.
\newblock


\bibitem[Alldieck et~al\mbox{.}(2019b)]%
        {alldieck2019tex2shape}
\bibfield{author}{\bibinfo{person}{Thiemo Alldieck}, \bibinfo{person}{Gerard
  Pons-Moll}, \bibinfo{person}{Christian Theobalt}, {and}
  \bibinfo{person}{Marcus Magnor}.} \bibinfo{year}{2019}\natexlab{b}.
\newblock \showarticletitle{Tex2shape: Detailed full human body geometry from a
  single image}. In \bibinfo{booktitle}{\emph{Proceedings of the IEEE/CVF
  International Conference on Computer Vision}}. \bibinfo{pages}{2293--2303}.
\newblock


\bibitem[Alldieck et~al\mbox{.}(2021)]%
        {Alldieck2021_imGHUM}
\bibfield{author}{\bibinfo{person}{Thiemo Alldieck}, \bibinfo{person}{Hongyi
  Xu}, {and} \bibinfo{person}{Cristian Sminchisescu}.}
  \bibinfo{year}{2021}\natexlab{}.
\newblock \showarticletitle{{imGHUM}: {I}mplicit Generative Models of 3D Human
  Shape and Articulated Pose}. In \bibinfo{booktitle}{\emph{International
  Conference on Computer Vision (ICCV)}}. \bibinfo{publisher}{{IEEE}},
  \bibinfo{pages}{5441--5450}.
\newblock


\bibitem[Anguelov et~al\mbox{.}(2005)]%
        {anguelov2005scape}
\bibfield{author}{\bibinfo{person}{Dragomir Anguelov}, \bibinfo{person}{Praveen
  Srinivasan}, \bibinfo{person}{Daphne Koller}, \bibinfo{person}{Sebastian
  Thrun}, \bibinfo{person}{Jim Rodgers}, {and} \bibinfo{person}{James Davis}.}
  \bibinfo{year}{2005}\natexlab{}.
\newblock \showarticletitle{{SCAPE}: {S}hape completion and animation of
  people}.
\newblock In \bibinfo{booktitle}{\emph{ACM SIGGRAPH 2005 Papers}}.
  \bibinfo{pages}{408--416}.
\newblock


\bibitem[Bertiche et~al\mbox{.}(2020)]%
        {bertiche2020cloth3d}
\bibfield{author}{\bibinfo{person}{Hugo Bertiche}, \bibinfo{person}{Meysam
  Madadi}, {and} \bibinfo{person}{Sergio Escalera}.}
  \bibinfo{year}{2020}\natexlab{}.
\newblock \showarticletitle{{CLOTH3D}: clothed 3d humans}. In
  \bibinfo{booktitle}{\emph{European Conference on Computer Vision (ECCV)}}.
  Springer, \bibinfo{pages}{344--359}.
\newblock


\bibitem[Chen et~al\mbox{.}(2021b)]%
        {chen2021animatable}
\bibfield{author}{\bibinfo{person}{Jianchuan Chen}, \bibinfo{person}{Ying
  Zhang}, \bibinfo{person}{Di Kang}, \bibinfo{person}{Xuefei Zhe},
  \bibinfo{person}{Linchao Bao}, \bibinfo{person}{Xu Jia}, {and}
  \bibinfo{person}{Huchuan Lu}.} \bibinfo{year}{2021}\natexlab{b}.
\newblock \bibinfo{title}{Animatable Neural Radiance Fields from Monocular RGB
  Videos}.
\newblock
\newblock
\showeprint[arxiv]{2106.13629}~[cs.CV]


\bibitem[Chen et~al\mbox{.}(2021a)]%
        {chen2021tightcap}
\bibfield{author}{\bibinfo{person}{Xin Chen}, \bibinfo{person}{Anqi Pang},
  \bibinfo{person}{Wei Yang}, \bibinfo{person}{Peihao Wang},
  \bibinfo{person}{Lan Xu}, {and} \bibinfo{person}{Jingyi Yu}.}
  \bibinfo{year}{2021}\natexlab{a}.
\newblock \showarticletitle{{TightCap}: {3D} Human Shape Capture with Clothing
  Tightness Field}.
\newblock \bibinfo{journal}{\emph{Transactions on Graphics (TOG)}}
  \bibinfo{volume}{41}, \bibinfo{number}{1} (\bibinfo{year}{2021}),
  \bibinfo{pages}{1--17}.
\newblock


\bibitem[Chibane et~al\mbox{.}(2020)]%
        {chibane2020ndf}
\bibfield{author}{\bibinfo{person}{Julian Chibane}, \bibinfo{person}{Aymen
  Mir}, {and} \bibinfo{person}{Gerard Pons-Moll}.}
  \bibinfo{year}{2020}\natexlab{}.
\newblock \showarticletitle{Neural Unsigned Distance Fields for Implicit
  Function Learning}. In \bibinfo{booktitle}{\emph{Advances in Neural
  Information Processing Systems (NeurIPS)}}.
\newblock


\bibitem[Choutas et~al\mbox{.}(2020)]%
        {Choutas2020ExPose}
\bibfield{author}{\bibinfo{person}{Vasileios Choutas},
  \bibinfo{person}{Georgios Pavlakos}, \bibinfo{person}{Timo Bolkart},
  \bibinfo{person}{Dimitrios Tzionas}, {and} \bibinfo{person}{Michael~J.
  Black}.} \bibinfo{year}{2020}\natexlab{}.
\newblock \showarticletitle{Monocular Expressive Body Regression through
  Body-Driven Attention}. In \bibinfo{booktitle}{\emph{European Conference on
  Computer Vision (ECCV)}}. \bibinfo{pages}{20--40}.
\newblock


\bibitem[Corona et~al\mbox{.}(2021)]%
        {corona2021smplicit}
\bibfield{author}{\bibinfo{person}{Enric Corona}, \bibinfo{person}{Albert
  Pumarola}, \bibinfo{person}{Guillem Alenya}, \bibinfo{person}{Gerard
  Pons-Moll}, {and} \bibinfo{person}{Francesc Moreno-Noguer}.}
  \bibinfo{year}{2021}\natexlab{}.
\newblock \showarticletitle{{SMPLicit}: {T}opology-aware generative model for
  clothed people}. In \bibinfo{booktitle}{\emph{Conference on Computer Vision
  and Pattern Recognition (CVPR)}}. \bibinfo{pages}{11875--11885}.
\newblock


\bibitem[Dabhi(2022)]%
        {clothsegmentation}
\bibfield{author}{\bibinfo{person}{Levin Dabhi}.}
  \bibinfo{year}{2022}\natexlab{}.
\newblock \bibinfo{title}{Clothes Segmentation using U2NET}.
\newblock
\newblock
\urldef\tempurl%
\url{https://github.com/levindabhi/cloth-segmentation}
\showURL{%
\tempurl}


\bibitem[Danecek et~al\mbox{.}(2022)]%
        {EMOCA:CVPR:2021}
\bibfield{author}{\bibinfo{person}{Radek Danecek}, \bibinfo{person}{Michael~J.
  Black}, {and} \bibinfo{person}{Timo Bolkart}.}
  \bibinfo{year}{2022}\natexlab{}.
\newblock \showarticletitle{{EMOCA}: {E}motion Driven Monocular Face Capture
  and Animation}. In \bibinfo{booktitle}{\emph{Conference on Computer Vision
  and Pattern Recognition (CVPR)}}.
\newblock


\bibitem[Feng et~al\mbox{.}(2021a)]%
        {PIXIE:3DV:2021}
\bibfield{author}{\bibinfo{person}{Yao Feng}, \bibinfo{person}{Vasileios
  Choutas}, \bibinfo{person}{Timo Bolkart}, \bibinfo{person}{Dimitrios
  Tzionas}, {and} \bibinfo{person}{Michael Black}.}
  \bibinfo{year}{2021}\natexlab{a}.
\newblock \showarticletitle{Collaborative Regression of Expressive Bodies using
  Moderation}. In \bibinfo{booktitle}{\emph{International Conference on 3D
  Vision (3DV)}}. \bibinfo{pages}{792--804}.
\newblock


\bibitem[Feng et~al\mbox{.}(2021b)]%
        {Feng:SIGGRAPH:2021}
\bibfield{author}{\bibinfo{person}{Yao Feng}, \bibinfo{person}{Haiwen Feng},
  \bibinfo{person}{Michael~J. Black}, {and} \bibinfo{person}{Timo Bolkart}.}
  \bibinfo{year}{2021}\natexlab{b}.
\newblock \showarticletitle{Learning an Animatable Detailed {3D} Face Model
  from In-the-Wild Images}.
\newblock \bibinfo{journal}{\emph{Transactions on Graphics, (Proc. SIGGRAPH)}}
  \bibinfo{volume}{40}, \bibinfo{number}{4} (\bibinfo{year}{2021}),
  \bibinfo{pages}{88:1--88:13}.
\newblock


\bibitem[Grassal et~al\mbox{.}(2022)]%
        {Grassal2022}
\bibfield{author}{\bibinfo{person}{Philip-William Grassal},
  \bibinfo{person}{Malte Prinzler}, \bibinfo{person}{Titus Leistner},
  \bibinfo{person}{Carsten Rother}, \bibinfo{person}{Matthias Nie{\ss}ner},
  {and} \bibinfo{person}{Justus Thies}.} \bibinfo{year}{2022}\natexlab{}.
\newblock \showarticletitle{Neural Head Avatars from Monocular {RGB} Videos}.
  In \bibinfo{booktitle}{\emph{Conference on Computer Vision and Pattern
  Recognition (CVPR)}}.
\newblock


\bibitem[He et~al\mbox{.}(2021)]%
        {he2021arch++}
\bibfield{author}{\bibinfo{person}{Tong He}, \bibinfo{person}{Yuanlu Xu},
  \bibinfo{person}{Shunsuke Saito}, \bibinfo{person}{Stefano Soatto}, {and}
  \bibinfo{person}{Tony Tung}.} \bibinfo{year}{2021}\natexlab{}.
\newblock \showarticletitle{{ARCH++}: {A}nimation-ready clothed human
  reconstruction revisited}. In \bibinfo{booktitle}{\emph{International
  Conference on Computer Vision (ICCV)}}. \bibinfo{pages}{11046--11056}.
\newblock


\bibitem[Hirshberg et~al\mbox{.}(2012)]%
        {Hirshberg2012_Coregistration}
\bibfield{author}{\bibinfo{person}{David~A. Hirshberg},
  \bibinfo{person}{Matthew Loper}, \bibinfo{person}{Eric Rachlin}, {and}
  \bibinfo{person}{Michael~J. Black}.} \bibinfo{year}{2012}\natexlab{}.
\newblock \showarticletitle{Coregistration: {S}imultaneous Alignment and
  Modeling of Articulated {3D} Shape}. In \bibinfo{booktitle}{\emph{European
  Conference on Computer Vision (ECCV)}} \emph{(\bibinfo{series}{Lecture Notes
  in Computer Science}, Vol.~\bibinfo{volume}{7577})}.
  \bibinfo{publisher}{Springer}, \bibinfo{pages}{242--255}.
\newblock


\bibitem[Hong et~al\mbox{.}(2022)]%
        {Hong2022headnerf}
\bibfield{author}{\bibinfo{person}{Yang Hong}, \bibinfo{person}{Bo Peng},
  \bibinfo{person}{Haiyao Xiao}, \bibinfo{person}{Ligang Liu}, {and}
  \bibinfo{person}{Juyong Zhang}.} \bibinfo{year}{2022}\natexlab{}.
\newblock \showarticletitle{{HeadNeRF}: {A} real-time nerf-based parametric
  head model}. In \bibinfo{booktitle}{\emph{Conference on Computer Vision and
  Pattern Recognition (CVPR)}}. \bibinfo{pages}{20374--20384}.
\newblock


\bibitem[Huang et~al\mbox{.}(2020)]%
        {huang2020arch}
\bibfield{author}{\bibinfo{person}{Zeng Huang}, \bibinfo{person}{Yuanlu Xu},
  \bibinfo{person}{Christoph Lassner}, \bibinfo{person}{Hao Li}, {and}
  \bibinfo{person}{Tony Tung}.} \bibinfo{year}{2020}\natexlab{}.
\newblock \showarticletitle{{ARCH}: {A}nimatable reconstruction of clothed
  humans}. In \bibinfo{booktitle}{\emph{Conference on Computer Vision and
  Pattern Recognition (CVPR)}}. \bibinfo{pages}{3093--3102}.
\newblock


\bibitem[Huber(1964)]%
        {Huber1964}
\bibfield{author}{\bibinfo{person}{Peter~J. Huber}.}
  \bibinfo{year}{1964}\natexlab{}.
\newblock \showarticletitle{{Robust Estimation of a Location Parameter}}.
\newblock \bibinfo{journal}{\emph{The Annals of Mathematical Statistics}}
  \bibinfo{volume}{35}, \bibinfo{number}{1} (\bibinfo{year}{1964}),
  \bibinfo{pages}{73 -- 101}.
\newblock


\bibitem[Jiang et~al\mbox{.}(2022)]%
        {jiang2022selfrecon}
\bibfield{author}{\bibinfo{person}{Boyi Jiang}, \bibinfo{person}{Yang Hong},
  \bibinfo{person}{Hujun Bao}, {and} \bibinfo{person}{Juyong Zhang}.}
  \bibinfo{year}{2022}\natexlab{}.
\newblock \showarticletitle{SelfRecon: Self Reconstruction Your Digital Avatar
  from Monocular Video}. In \bibinfo{booktitle}{\emph{Proceedings of the
  IEEE/CVF Conference on Computer Vision and Pattern Recognition}}.
  \bibinfo{pages}{5605--5615}.
\newblock


\bibitem[Jiang et~al\mbox{.}(2020)]%
        {jiang2020bcnet}
\bibfield{author}{\bibinfo{person}{Boyi Jiang}, \bibinfo{person}{Juyong Zhang},
  \bibinfo{person}{Yang Hong}, \bibinfo{person}{Jinhao Luo},
  \bibinfo{person}{Ligang Liu}, {and} \bibinfo{person}{Hujun Bao}.}
  \bibinfo{year}{2020}\natexlab{}.
\newblock \showarticletitle{{BCNet}: {L}earning body and cloth shape from a
  single image}. In \bibinfo{booktitle}{\emph{European Conference on Computer
  Vision (ECCV)}}. Springer, \bibinfo{pages}{18--35}.
\newblock


\bibitem[Jin et~al\mbox{.}(2020)]%
        {jin2020pixel}
\bibfield{author}{\bibinfo{person}{Ning Jin}, \bibinfo{person}{Yilin Zhu},
  \bibinfo{person}{Zhenglin Geng}, {and} \bibinfo{person}{Ronald Fedkiw}.}
  \bibinfo{year}{2020}\natexlab{}.
\newblock \showarticletitle{A Pixel-Based Framework for Data-Driven Clothing}.
  In \bibinfo{booktitle}{\emph{Computer Graphics Forum}},
  Vol.~\bibinfo{volume}{39}. Wiley Online Library, \bibinfo{pages}{135--144}.
\newblock


\bibitem[Joo et~al\mbox{.}(2018)]%
        {Joo2018Adam}
\bibfield{author}{\bibinfo{person}{Hanbyul Joo}, \bibinfo{person}{Tomas Simon},
  {and} \bibinfo{person}{Yaser Sheikh}.} \bibinfo{year}{2018}\natexlab{}.
\newblock \showarticletitle{Total Capture: {A} 3D Deformation Model for
  Tracking Faces, Hands, and Bodies}. In \bibinfo{booktitle}{\emph{Conference
  on Computer Vision and Pattern Recognition (CVPR)}}.
  \bibinfo{pages}{8320--8329}.
\newblock


\bibitem[Kanazawa et~al\mbox{.}(2018)]%
        {Kanazawa2018_hmr}
\bibfield{author}{\bibinfo{person}{Angjoo Kanazawa},
  \bibinfo{person}{Michael~J. Black}, \bibinfo{person}{David~W. Jacobs}, {and}
  \bibinfo{person}{Jitendra Malik}.} \bibinfo{year}{2018}\natexlab{}.
\newblock \showarticletitle{End-to-end Recovery of Human Shape and Pose}. In
  \bibinfo{booktitle}{\emph{Conference on Computer Vision and Pattern
  Recognition (CVPR)}}. \bibinfo{pages}{7122--7131}.
\newblock


\bibitem[Kingma and Ba(2015)]%
        {Kingma2015}
\bibfield{author}{\bibinfo{person}{Diederik~P. Kingma} {and}
  \bibinfo{person}{Jimmy Ba}.} \bibinfo{year}{2015}\natexlab{}.
\newblock \showarticletitle{Adam: {A} Method for Stochastic Optimization}. In
  \bibinfo{booktitle}{\emph{International Conference on Learning
  Representations (ICLR)}}.
\newblock


\bibitem[Kolotouros et~al\mbox{.}(2019)]%
        {Kolotouros2019_SPIN}
\bibfield{author}{\bibinfo{person}{Nikos Kolotouros}, \bibinfo{person}{Georgios
  Pavlakos}, \bibinfo{person}{Michael~J. Black}, {and} \bibinfo{person}{Kostas
  Daniilidis}.} \bibinfo{year}{2019}\natexlab{}.
\newblock \showarticletitle{Learning to Reconstruct {3D} Human Pose and Shape
  via Model-Fitting in the Loop}. In \bibinfo{booktitle}{\emph{International
  Conference on Computer Vision (ICCV)}}. \bibinfo{publisher}{{IEEE}},
  \bibinfo{pages}{2252--2261}.
\newblock


\bibitem[Lazova et~al\mbox{.}(2019)]%
        {lazova3dv2019}
\bibfield{author}{\bibinfo{person}{Verica Lazova}, \bibinfo{person}{Eldar
  Insafutdinov}, {and} \bibinfo{person}{Gerard Pons-Moll}.}
  \bibinfo{year}{2019}\natexlab{}.
\newblock \showarticletitle{360-Degree Textures of People in Clothing from a
  Single Image}. In \bibinfo{booktitle}{\emph{International Conference on 3D
  Vision (3DV)}}.
\newblock


\bibitem[Lin et~al\mbox{.}(2022)]%
        {RobustVideoMatting}
\bibfield{author}{\bibinfo{person}{Shanchuan Lin}, \bibinfo{person}{Linjie
  Yang}, \bibinfo{person}{Imran Saleemi}, {and} \bibinfo{person}{Soumyadip
  Sengupta}.} \bibinfo{year}{2022}\natexlab{}.
\newblock \bibinfo{title}{Robust Video Matting (RVM)}.
\newblock
\newblock
\urldef\tempurl%
\url{https://github.com/PeterL1n/RobustVideoMatting}
\showURL{%
\tempurl}


\bibitem[Liu et~al\mbox{.}(2021b)]%
        {liu2021neural}
\bibfield{author}{\bibinfo{person}{Lingjie Liu}, \bibinfo{person}{Marc
  Habermann}, \bibinfo{person}{Viktor Rudnev}, \bibinfo{person}{Kripasindhu
  Sarkar}, \bibinfo{person}{Jiatao Gu}, {and} \bibinfo{person}{Christian
  Theobalt}.} \bibinfo{year}{2021}\natexlab{b}.
\newblock \showarticletitle{Neural actor: {N}eural free-view synthesis of human
  actors with pose control}.
\newblock \bibinfo{journal}{\emph{Transactions on Graphics (TOG)}}
  \bibinfo{volume}{40}, \bibinfo{number}{6} (\bibinfo{year}{2021}),
  \bibinfo{pages}{1--16}.
\newblock


\bibitem[Liu et~al\mbox{.}(2020)]%
        {liu2020dist}
\bibfield{author}{\bibinfo{person}{Shaohui Liu}, \bibinfo{person}{Yinda Zhang},
  \bibinfo{person}{Songyou Peng}, \bibinfo{person}{Boxin Shi},
  \bibinfo{person}{Marc Pollefeys}, {and} \bibinfo{person}{Zhaopeng Cui}.}
  \bibinfo{year}{2020}\natexlab{}.
\newblock \showarticletitle{Dist: Rendering deep implicit signed distance
  function with differentiable sphere tracing}. In
  \bibinfo{booktitle}{\emph{Conference on Computer Vision and Pattern
  Recognition (CVPR)}}. \bibinfo{pages}{2019--2028}.
\newblock


\bibitem[Liu et~al\mbox{.}(2021a)]%
        {Liu2021hmrsurvey}
\bibfield{author}{\bibinfo{person}{Wu Liu}, \bibinfo{person}{Qian Bao},
  \bibinfo{person}{Yu Sun}, {and} \bibinfo{person}{Tao Mei}.}
  \bibinfo{year}{2021}\natexlab{a}.
\newblock \showarticletitle{Recent Advances in Monocular {2D} and {3D} Human
  Pose Estimation: {A} Deep Learning Perspective}.
\newblock \bibinfo{journal}{\emph{CoRR}}  \bibinfo{volume}{abs/2104.11536}
  (\bibinfo{year}{2021}).
\newblock


\bibitem[Liu et~al\mbox{.}(2019)]%
        {lwb2019}
\bibfield{author}{\bibinfo{person}{Wen Liu}, \bibinfo{person}{Zhixin Piao},
  \bibinfo{person}{Min Jie}, \bibinfo{person}{Wenhan Luo}, \bibinfo{person}{Lin
  Ma}, {and} \bibinfo{person}{Shenghua Gao}.} \bibinfo{year}{2019}\natexlab{}.
\newblock \showarticletitle{Liquid Warping {GAN}: {A} Unified Framework for
  Human Motion Imitation, Appearance Transfer and Novel View Synthesis}. In
  \bibinfo{booktitle}{\emph{International Conference on Computer Vision
  (ICCV)}}. \bibinfo{pages}{5903--5912}.
\newblock


\bibitem[Loper et~al\mbox{.}(2015)]%
        {SMPL:2015}
\bibfield{author}{\bibinfo{person}{Matthew Loper}, \bibinfo{person}{Naureen
  Mahmood}, \bibinfo{person}{Javier Romero}, \bibinfo{person}{Gerard
  Pons-Moll}, {and} \bibinfo{person}{Michael~J. Black}.}
  \bibinfo{year}{2015}\natexlab{}.
\newblock \showarticletitle{{SMPL}: {A} Skinned Multi-Person Linear Model}.
\newblock \bibinfo{journal}{\emph{Transactions on Graphics, (Proc. SIGGRAPH
  Asia)}} \bibinfo{volume}{34}, \bibinfo{number}{6} (\bibinfo{year}{2015}),
  \bibinfo{pages}{248:1--248:16}.
\newblock


\bibitem[Ma et~al\mbox{.}(2020a)]%
        {ma2020learning}
\bibfield{author}{\bibinfo{person}{Qianli Ma}, \bibinfo{person}{Jinlong Yang},
  \bibinfo{person}{Anurag Ranjan}, \bibinfo{person}{Sergi Pujades},
  \bibinfo{person}{Gerard Pons-Moll}, \bibinfo{person}{Siyu Tang}, {and}
  \bibinfo{person}{Michael~J Black}.} \bibinfo{year}{2020}\natexlab{a}.
\newblock \showarticletitle{Learning to dress 3{D} people in generative
  clothing}. In \bibinfo{booktitle}{\emph{Conference on Computer Vision and
  Pattern Recognition (CVPR)}}. \bibinfo{pages}{6469--6478}.
\newblock


\bibitem[Ma et~al\mbox{.}(2020b)]%
        {CAPE:CVPR:20}
\bibfield{author}{\bibinfo{person}{Qianli Ma}, \bibinfo{person}{Jinlong Yang},
  \bibinfo{person}{Anurag Ranjan}, \bibinfo{person}{Sergi Pujades},
  \bibinfo{person}{Gerard Pons-Moll}, \bibinfo{person}{Siyu Tang}, {and}
  \bibinfo{person}{Michael~J. Black}.} \bibinfo{year}{2020}\natexlab{b}.
\newblock \showarticletitle{Learning to Dress {3D} People in Generative
  Clothing}. In \bibinfo{booktitle}{\emph{Conference on Computer Vision and
  Pattern Recognition (CVPR)}}. \bibinfo{pages}{6468--6477}.
\newblock


\bibitem[Mihajlovic et~al\mbox{.}(2021)]%
        {LEAP:CVPR:2021}
\bibfield{author}{\bibinfo{person}{Marko Mihajlovic}, \bibinfo{person}{Yan
  Zhang}, \bibinfo{person}{Michael~J. Black}, {and} \bibinfo{person}{Siyu
  Tang}.} \bibinfo{year}{2021}\natexlab{}.
\newblock \showarticletitle{{LEAP}: Learning Articulated Occupancy of People}.
  In \bibinfo{booktitle}{\emph{Proceedings IEEE/CVF Conf. on Computer Vision
  and Pattern Recognition (CVPR)}}. \bibinfo{pages}{10461--1047}.
\newblock


\bibitem[Mildenhall et~al\mbox{.}(2020)]%
        {mildenhall2020nerf}
\bibfield{author}{\bibinfo{person}{Ben Mildenhall}, \bibinfo{person}{Pratul~P
  Srinivasan}, \bibinfo{person}{Matthew Tancik}, \bibinfo{person}{Jonathan~T
  Barron}, \bibinfo{person}{Ravi Ramamoorthi}, {and} \bibinfo{person}{Ren Ng}.}
  \bibinfo{year}{2020}\natexlab{}.
\newblock \showarticletitle{Nerf: Representing scenes as neural radiance fields
  for view synthesis}. In \bibinfo{booktitle}{\emph{European Conference on
  Computer Vision (ECCV)}}. Springer, \bibinfo{pages}{405--421}.
\newblock


\bibitem[Niemeyer et~al\mbox{.}(2020)]%
        {niemeyer2020differentiable}
\bibfield{author}{\bibinfo{person}{Michael Niemeyer}, \bibinfo{person}{Lars
  Mescheder}, \bibinfo{person}{Michael Oechsle}, {and} \bibinfo{person}{Andreas
  Geiger}.} \bibinfo{year}{2020}\natexlab{}.
\newblock \showarticletitle{Differentiable volumetric rendering: Learning
  implicit 3d representations without 3d supervision}. In
  \bibinfo{booktitle}{\emph{Conference on Computer Vision and Pattern
  Recognition (CVPR)}}. \bibinfo{pages}{3504--3515}.
\newblock


\bibitem[Omran et~al\mbox{.}(2018)]%
        {omran2018neural}
\bibfield{author}{\bibinfo{person}{Mohamed Omran}, \bibinfo{person}{Christoph
  Lassner}, \bibinfo{person}{Gerard Pons-Moll}, \bibinfo{person}{Peter Gehler},
  {and} \bibinfo{person}{Bernt Schiele}.} \bibinfo{year}{2018}\natexlab{}.
\newblock \showarticletitle{Neural body fitting: Unifying deep learning and
  model based human pose and shape estimation}. In
  \bibinfo{booktitle}{\emph{International Conference on Computer Vision
  (ICCV)}}. IEEE, \bibinfo{pages}{484--494}.
\newblock


\bibitem[Osman et~al\mbox{.}(2020)]%
        {osman2020star}
\bibfield{author}{\bibinfo{person}{Ahmed A.~A. Osman}, \bibinfo{person}{Timo
  Bolkart}, {and} \bibinfo{person}{Michael~J. Black}.}
  \bibinfo{year}{2020}\natexlab{}.
\newblock \showarticletitle{{STAR}: {S}parse trained articulated human body
  regressor}. In \bibinfo{booktitle}{\emph{European Conference on Computer
  Vision (ECCV)}}. \bibinfo{pages}{598--613}.
\newblock


\bibitem[Patel et~al\mbox{.}(2020)]%
        {patel2020tailornet}
\bibfield{author}{\bibinfo{person}{Chaitanya Patel},
  \bibinfo{person}{Zhouyingcheng Liao}, {and} \bibinfo{person}{Gerard
  Pons-Moll}.} \bibinfo{year}{2020}\natexlab{}.
\newblock \showarticletitle{Tailornet: {P}redicting clothing in 3d as a
  function of human pose, shape and garment style}. In
  \bibinfo{booktitle}{\emph{Conference on Computer Vision and Pattern
  Recognition (CVPR)}}. \bibinfo{pages}{7365--7375}.
\newblock


\bibitem[Pavlakos et~al\mbox{.}(2019)]%
        {Pavlakos2019_smplifyx}
\bibfield{author}{\bibinfo{person}{Georgios Pavlakos},
  \bibinfo{person}{Vasileios Choutas}, \bibinfo{person}{Nima Ghorbani},
  \bibinfo{person}{Timo Bolkart}, \bibinfo{person}{Ahmed A.~A. Osman},
  \bibinfo{person}{Dimitrios Tzionas}, {and} \bibinfo{person}{Michael~J.
  Black}.} \bibinfo{year}{2019}\natexlab{}.
\newblock \showarticletitle{Expressive Body Capture: {3D} Hands, Face, and Body
  From a Single Image}. In \bibinfo{booktitle}{\emph{Conference on Computer
  Vision and Pattern Recognition (CVPR)}}. \bibinfo{pages}{10975--10985}.
\newblock


\bibitem[Peng et~al\mbox{.}(2021a)]%
        {peng2021animatable}
\bibfield{author}{\bibinfo{person}{Sida Peng}, \bibinfo{person}{Junting Dong},
  \bibinfo{person}{Qianqian Wang}, \bibinfo{person}{Shangzhan Zhang},
  \bibinfo{person}{Qing Shuai}, \bibinfo{person}{Xiaowei Zhou}, {and}
  \bibinfo{person}{Hujun Bao}.} \bibinfo{year}{2021}\natexlab{a}.
\newblock \showarticletitle{Animatable Neural Radiance Fields for Modeling
  Dynamic Human Bodies}. In \bibinfo{booktitle}{\emph{ICCV}}.
\newblock


\bibitem[Peng et~al\mbox{.}(2022)]%
        {peng2022animatable}
\bibfield{author}{\bibinfo{person}{Sida Peng}, \bibinfo{person}{Shangzhan
  Zhang}, \bibinfo{person}{Zhen Xu}, \bibinfo{person}{Chen Geng},
  \bibinfo{person}{Boyi Jiang}, \bibinfo{person}{Hujun Bao}, {and}
  \bibinfo{person}{Xiaowei Zhou}.} \bibinfo{year}{2022}\natexlab{}.
\newblock \showarticletitle{Animatable Neural Implicit Surfaces for Creating
  Avatars from Videos}.
\newblock \bibinfo{journal}{\emph{arXiv preprint arXiv:2203.08133}}
  (\bibinfo{year}{2022}).
\newblock


\bibitem[Peng et~al\mbox{.}(2021b)]%
        {peng2021neural}
\bibfield{author}{\bibinfo{person}{Sida Peng}, \bibinfo{person}{Yuanqing
  Zhang}, \bibinfo{person}{Yinghao Xu}, \bibinfo{person}{Qianqian Wang},
  \bibinfo{person}{Qing Shuai}, \bibinfo{person}{Hujun Bao}, {and}
  \bibinfo{person}{Xiaowei Zhou}.} \bibinfo{year}{2021}\natexlab{b}.
\newblock \showarticletitle{Neural body: {I}mplicit neural representations with
  structured latent codes for novel view synthesis of dynamic humans}. In
  \bibinfo{booktitle}{\emph{Conference on Computer Vision and Pattern
  Recognition (CVPR)}}. \bibinfo{pages}{9054--9063}.
\newblock


\bibitem[Pons-Moll et~al\mbox{.}(2017)]%
        {pons2017clothcap}
\bibfield{author}{\bibinfo{person}{Gerard Pons-Moll}, \bibinfo{person}{Sergi
  Pujades}, \bibinfo{person}{Sonny Hu}, {and} \bibinfo{person}{Michael~J
  Black}.} \bibinfo{year}{2017}\natexlab{}.
\newblock \showarticletitle{{ClothCap}: {S}eamless {4D} clothing capture and
  retargeting}.
\newblock \bibinfo{journal}{\emph{Transactions on Graphics (TOG)}}
  \bibinfo{volume}{36}, \bibinfo{number}{4} (\bibinfo{year}{2017}),
  \bibinfo{pages}{1--15}.
\newblock


\bibitem[Prokudin et~al\mbox{.}(2021)]%
        {SMPLpix:WACV:2020}
\bibfield{author}{\bibinfo{person}{Sergey Prokudin},
  \bibinfo{person}{Michael~J. Black}, {and} \bibinfo{person}{Javier Romero}.}
  \bibinfo{year}{2021}\natexlab{}.
\newblock \showarticletitle{{SMPLpix}: Neural Avatars from {3D} Human Models}.
  In \bibinfo{booktitle}{\emph{Winter Conference on Applications of Computer
  Vision (WACV)}}. \bibinfo{pages}{1810--1819}.
\newblock


\bibitem[Ravi et~al\mbox{.}(2020)]%
        {Ravi2020_PyTorch3D}
\bibfield{author}{\bibinfo{person}{Nikhila Ravi}, \bibinfo{person}{Jeremy
  Reizenstein}, \bibinfo{person}{David Novotny}, \bibinfo{person}{Taylor
  Gordon}, \bibinfo{person}{Wan-Yen Lo}, \bibinfo{person}{Justin Johnson},
  {and} \bibinfo{person}{Georgia Gkioxari}.} \bibinfo{year}{2020}\natexlab{}.
\newblock \showarticletitle{Accelerating {3D} Deep Learning with PyTorch3D}.
\newblock \bibinfo{journal}{\emph{arXiv:2007.08501}} (\bibinfo{year}{2020}).
\newblock


\bibitem[Rong et~al\mbox{.}(2021)]%
        {Rong2021Frankmocap}
\bibfield{author}{\bibinfo{person}{Yu Rong}, \bibinfo{person}{Takaaki
  Shiratori}, {and} \bibinfo{person}{Hanbyul Joo}.}
  \bibinfo{year}{2021}\natexlab{}.
\newblock \showarticletitle{{FrankMocap}: {A} Monocular {3D} Whole-Body Pose
  Estimation System via Regression and Integration}. In
  \bibinfo{booktitle}{\emph{International Conference on Computer Vision
  Workshops (ICCV-W)}}.
\newblock


\bibitem[Saito et~al\mbox{.}(2019)]%
        {saito2019pifu}
\bibfield{author}{\bibinfo{person}{Shunsuke Saito}, \bibinfo{person}{Zeng
  Huang}, \bibinfo{person}{Ryota Natsume}, \bibinfo{person}{Shigeo Morishima},
  \bibinfo{person}{Angjoo Kanazawa}, {and} \bibinfo{person}{Hao Li}.}
  \bibinfo{year}{2019}\natexlab{}.
\newblock \showarticletitle{{PIFu}: {P}ixel-Aligned Implicit Function for
  High-Resolution Clothed Human Digitization}. In
  \bibinfo{booktitle}{\emph{International Conference on Computer Vision
  (ICCV)}}.
\newblock


\bibitem[Saito et~al\mbox{.}(2020)]%
        {saito2020pifuhd}
\bibfield{author}{\bibinfo{person}{Shunsuke Saito}, \bibinfo{person}{Tomas
  Simon}, \bibinfo{person}{Jason Saragih}, {and} \bibinfo{person}{Hanbyul
  Joo}.} \bibinfo{year}{2020}\natexlab{}.
\newblock \showarticletitle{{PIFuHD}: {M}ulti-Level Pixel-Aligned Implicit
  Function for High-Resolution 3D Human Digitization}. In
  \bibinfo{booktitle}{\emph{Conference on Computer Vision and Pattern
  Recognition (CVPR)}}.
\newblock


\bibitem[Santesteban et~al\mbox{.}(2019)]%
        {santesteban2019learning}
\bibfield{author}{\bibinfo{person}{Igor Santesteban}, \bibinfo{person}{Miguel~A
  Otaduy}, {and} \bibinfo{person}{Dan Casas}.} \bibinfo{year}{2019}\natexlab{}.
\newblock \showarticletitle{Learning-based animation of clothing for virtual
  try-on}. In \bibinfo{booktitle}{\emph{Computer Graphics Forum}},
  Vol.~\bibinfo{volume}{38}. Wiley Online Library, \bibinfo{pages}{355--366}.
\newblock


\bibitem[Su et~al\mbox{.}(2021)]%
        {su2021nerf}
\bibfield{author}{\bibinfo{person}{Shih-Yang Su}, \bibinfo{person}{Frank Yu},
  \bibinfo{person}{Michael Zollh{\"o}fer}, {and} \bibinfo{person}{Helge
  Rhodin}.} \bibinfo{year}{2021}\natexlab{}.
\newblock \showarticletitle{{A-NeRF}: {A}rticulated neural radiance fields for
  learning human shape, appearance, and pose}.
\newblock \bibinfo{journal}{\emph{Advances in Neural Information Processing
  Systems (NeurIPS)}}  \bibinfo{volume}{34} (\bibinfo{year}{2021}).
\newblock


\bibitem[Tian et~al\mbox{.}(2022)]%
        {tian2022hmrsurvey}
\bibfield{author}{\bibinfo{person}{Yating Tian}, \bibinfo{person}{Hongwen
  Zhang}, \bibinfo{person}{Yebin Liu}, {and} \bibinfo{person}{Limin Wang}.}
  \bibinfo{year}{2022}\natexlab{}.
\newblock \showarticletitle{Recovering {3D} Human Mesh from Monocular Images:
  {A} Survey}.
\newblock \bibinfo{journal}{\emph{arXiv preprint arXiv:2203.01923}}
  (\bibinfo{year}{2022}).
\newblock


\bibitem[Tiwari et~al\mbox{.}(2020)]%
        {tiwari2020sizer}
\bibfield{author}{\bibinfo{person}{Garvita Tiwari}, \bibinfo{person}{Bharat~Lal
  Bhatnagar}, \bibinfo{person}{Tony Tung}, {and} \bibinfo{person}{Gerard
  Pons-Moll}.} \bibinfo{year}{2020}\natexlab{}.
\newblock \showarticletitle{{SIZER}: {A} dataset and model for parsing 3d
  clothing and learning size sensitive 3d clothing}. In
  \bibinfo{booktitle}{\emph{European Conference on Computer Vision (ECCV)}}.
  Springer, \bibinfo{pages}{1--18}.
\newblock


\bibitem[Vidaurre et~al\mbox{.}(2020)]%
        {vidaurre2020fully}
\bibfield{author}{\bibinfo{person}{Raquel Vidaurre}, \bibinfo{person}{Igor
  Santesteban}, \bibinfo{person}{Elena Garces}, {and} \bibinfo{person}{Dan
  Casas}.} \bibinfo{year}{2020}\natexlab{}.
\newblock \showarticletitle{Fully Convolutional Graph Neural Networks for
  Parametric Virtual Try-On}. In \bibinfo{booktitle}{\emph{Computer Graphics
  Forum}}, Vol.~\bibinfo{volume}{39}. Wiley Online Library,
  \bibinfo{pages}{145--156}.
\newblock


\bibitem[Wang et~al\mbox{.}(2018)]%
        {Wang2018}
\bibfield{author}{\bibinfo{person}{Yi Wang}, \bibinfo{person}{Xin Tao},
  \bibinfo{person}{Xiaojuan Qi}, \bibinfo{person}{Xiaoyong Shen}, {and}
  \bibinfo{person}{Jiaya Jia}.} \bibinfo{year}{2018}\natexlab{}.
\newblock \showarticletitle{Image inpainting via generative multi-column
  convolutional neural networks}. In \bibinfo{booktitle}{\emph{Advances in
  Neural Information Processing Systems (NeurIPS)}}. \bibinfo{pages}{331--340}.
\newblock


\bibitem[Weng et~al\mbox{.}(2022)]%
        {weng2022humannerf}
\bibfield{author}{\bibinfo{person}{Chung-Yi Weng}, \bibinfo{person}{Brian
  Curless}, \bibinfo{person}{Pratul~P. Srinivasan},
  \bibinfo{person}{Jonathan~T. Barron}, {and} \bibinfo{person}{Ira
  Kemelmacher-Shlizerman}.} \bibinfo{year}{2022}\natexlab{}.
\newblock \showarticletitle{Human{N}e{RF}: Free-Viewpoint Rendering of Moving
  People From Monocular Video}. In \bibinfo{booktitle}{\emph{Conference on
  Computer Vision and Pattern Recognition (CVPR)}}.
  \bibinfo{pages}{16210--16220}.
\newblock


\bibitem[Xiang et~al\mbox{.}(2019)]%
        {Xiang2019}
\bibfield{author}{\bibinfo{person}{Donglai Xiang}, \bibinfo{person}{Hanbyul
  Joo}, {and} \bibinfo{person}{Yaser Sheikh}.} \bibinfo{year}{2019}\natexlab{}.
\newblock \showarticletitle{Monocular Total Capture: {P}osing Face, Body, and
  Hands in the Wild}. In \bibinfo{booktitle}{\emph{Conference on Computer
  Vision and Pattern Recognition (CVPR)}}. \bibinfo{pages}{10965--10974}.
\newblock


\bibitem[Xiang et~al\mbox{.}(2021)]%
        {xiang2021modeling}
\bibfield{author}{\bibinfo{person}{Donglai Xiang}, \bibinfo{person}{Fabian
  Prada}, \bibinfo{person}{Timur Bagautdinov}, \bibinfo{person}{Weipeng Xu},
  \bibinfo{person}{Yuan Dong}, \bibinfo{person}{He Wen},
  \bibinfo{person}{Jessica Hodgins}, {and} \bibinfo{person}{Chenglei Wu}.}
  \bibinfo{year}{2021}\natexlab{}.
\newblock \showarticletitle{Modeling clothing as a separate layer for an
  animatable human avatar}.
\newblock \bibinfo{journal}{\emph{Transactions on Graphics (TOG)}}
  \bibinfo{volume}{40}, \bibinfo{number}{6} (\bibinfo{year}{2021}),
  \bibinfo{pages}{1--15}.
\newblock


\bibitem[Xiu et~al\mbox{.}(2022)]%
        {xiu2022icon}
\bibfield{author}{\bibinfo{person}{Yuliang Xiu}, \bibinfo{person}{Jinlong
  Yang}, \bibinfo{person}{Dimitrios Tzionas}, {and} \bibinfo{person}{Michael~J.
  Black}.} \bibinfo{year}{2022}\natexlab{}.
\newblock \showarticletitle{{ICON}: {I}mplicit {C}lothed humans {O}btained from
  {N}ormals}. In \bibinfo{booktitle}{\emph{Conference on Computer Vision and
  Pattern Recognition (CVPR)}}.
\newblock


\bibitem[Xu et~al\mbox{.}(2021)]%
        {xu2021h}
\bibfield{author}{\bibinfo{person}{Hongyi Xu}, \bibinfo{person}{Thiemo
  Alldieck}, {and} \bibinfo{person}{Cristian Sminchisescu}.}
  \bibinfo{year}{2021}\natexlab{}.
\newblock \showarticletitle{{H-NeRF}: {N}eural radiance fields for rendering
  and temporal reconstruction of humans in motion}.
\newblock \bibinfo{journal}{\emph{Advances in Neural Information Processing
  Systems (NeurIPS)}}  \bibinfo{volume}{34} (\bibinfo{year}{2021}).
\newblock


\bibitem[Xu et~al\mbox{.}(2020)]%
        {xu2020ghum}
\bibfield{author}{\bibinfo{person}{Hongyi Xu}, \bibinfo{person}{Eduard~Gabriel
  Bazavan}, \bibinfo{person}{Andrei Zanfir}, \bibinfo{person}{William~T
  Freeman}, \bibinfo{person}{Rahul Sukthankar}, {and} \bibinfo{person}{Cristian
  Sminchisescu}.} \bibinfo{year}{2020}\natexlab{}.
\newblock \showarticletitle{{GHUM \& GHUML}: {G}enerative 3d human shape and
  articulated pose models}. In \bibinfo{booktitle}{\emph{Conference on Computer
  Vision and Pattern Recognition (CVPR)}}. \bibinfo{pages}{6184--6193}.
\newblock


\bibitem[Yang et~al\mbox{.}(2020)]%
        {Yang2020PRRCNN}
\bibfield{author}{\bibinfo{person}{Lu Yang}, \bibinfo{person}{Qing Song},
  \bibinfo{person}{Zhihui Wang}, \bibinfo{person}{Mengjie Hu},
  \bibinfo{person}{Chun Liu}, \bibinfo{person}{Xueshi Xin},
  \bibinfo{person}{Wenhe Jia}, {and} \bibinfo{person}{Songcen Xu}.}
  \bibinfo{year}{2020}\natexlab{}.
\newblock \showarticletitle{Renovating Parsing {R-CNN} for Accurate Multiple
  Human Parsing}. In \bibinfo{booktitle}{\emph{European Conference on Computer
  Vision (ECCV)}} \emph{(\bibinfo{series}{Lecture Notes in Computer Science},
  Vol.~\bibinfo{volume}{12357})}. \bibinfo{publisher}{Springer},
  \bibinfo{pages}{421--437}.
\newblock


\bibitem[Yang et~al\mbox{.}(2021)]%
        {yang2021s3}
\bibfield{author}{\bibinfo{person}{Ze Yang}, \bibinfo{person}{Shenlong Wang},
  \bibinfo{person}{Sivabalan Manivasagam}, \bibinfo{person}{Zeng Huang},
  \bibinfo{person}{Wei-Chiu Ma}, \bibinfo{person}{Xinchen Yan},
  \bibinfo{person}{Ersin Yumer}, {and} \bibinfo{person}{Raquel Urtasun}.}
  \bibinfo{year}{2021}\natexlab{}.
\newblock \showarticletitle{{S3}: {N}eural shape, skeleton, and skinning fields
  for {3D} human modeling}. In \bibinfo{booktitle}{\emph{Conference on Computer
  Vision and Pattern Recognition (CVPR)}}. \bibinfo{pages}{13284--13293}.
\newblock


\bibitem[Yariv et~al\mbox{.}(2021)]%
        {yariv2021volume}
\bibfield{author}{\bibinfo{person}{Lior Yariv}, \bibinfo{person}{Jiatao Gu},
  \bibinfo{person}{Yoni Kasten}, {and} \bibinfo{person}{Yaron Lipman}.}
  \bibinfo{year}{2021}\natexlab{}.
\newblock \showarticletitle{Volume rendering of neural implicit surfaces}.
\newblock \bibinfo{journal}{\emph{Advances in Neural Information Processing
  Systems (NeurIPS)}}  \bibinfo{volume}{34} (\bibinfo{year}{2021}).
\newblock


\bibitem[Yariv et~al\mbox{.}(2020)]%
        {yariv2020multiview}
\bibfield{author}{\bibinfo{person}{Lior Yariv}, \bibinfo{person}{Yoni Kasten},
  \bibinfo{person}{Dror Moran}, \bibinfo{person}{Meirav Galun},
  \bibinfo{person}{Matan Atzmon}, \bibinfo{person}{Basri Ronen}, {and}
  \bibinfo{person}{Yaron Lipman}.} \bibinfo{year}{2020}\natexlab{}.
\newblock \showarticletitle{Multiview neural surface reconstruction by
  disentangling geometry and appearance}.
\newblock \bibinfo{journal}{\emph{Advances in Neural Information Processing
  Systems (NeurIPS)}}  \bibinfo{volume}{33} (\bibinfo{year}{2020}),
  \bibinfo{pages}{2492--2502}.
\newblock


\bibitem[Zanfir et~al\mbox{.}(2021)]%
        {Zanfir2021_HUND}
\bibfield{author}{\bibinfo{person}{Andrei Zanfir},
  \bibinfo{person}{Eduard~Gabriel Bazavan}, \bibinfo{person}{Mihai Zanfir},
  \bibinfo{person}{William~T. Freeman}, \bibinfo{person}{Rahul Sukthankar},
  {and} \bibinfo{person}{Cristian Sminchisescu}.}
  \bibinfo{year}{2021}\natexlab{}.
\newblock \showarticletitle{Neural Descent for Visual {3D} Human Pose and
  Shape}. In \bibinfo{booktitle}{\emph{Conference on Computer Vision and
  Pattern Recognition (CVPR)}}. \bibinfo{publisher}{Computer Vision Foundation
  / {IEEE}}, \bibinfo{pages}{14484--14493}.
\newblock


\bibitem[Zheng et~al\mbox{.}(2021)]%
        {zheng2021pamir}
\bibfield{author}{\bibinfo{person}{Zerong Zheng}, \bibinfo{person}{Tao Yu},
  \bibinfo{person}{Yebin Liu}, {and} \bibinfo{person}{Qionghai Dai}.}
  \bibinfo{year}{2021}\natexlab{}.
\newblock \showarticletitle{{PaMIR}: {P}arametric model-conditioned implicit
  representation for image-based human reconstruction}.
\newblock \bibinfo{journal}{\emph{Transactions on Pattern Analysis and Machine
  Intelligence (PAMI)}} (\bibinfo{year}{2021}).
\newblock


\bibitem[Zhou et~al\mbox{.}(2021)]%
        {Zhou2021}
\bibfield{author}{\bibinfo{person}{Yuxiao Zhou}, \bibinfo{person}{Marc
  Habermann}, \bibinfo{person}{Ikhsanul Habibie}, \bibinfo{person}{Ayush
  Tewari}, \bibinfo{person}{Christian Theobalt}, {and} \bibinfo{person}{Feng
  Xu}.} \bibinfo{year}{2021}\natexlab{}.
\newblock \showarticletitle{Monocular Real-Time Full Body Capture With
  Inter-Part Correlations}. In \bibinfo{booktitle}{\emph{Conference on Computer
  Vision and Pattern Recognition (CVPR)}}. \bibinfo{pages}{4811--4822}.
\newblock


\bibitem[Zhu et~al\mbox{.}(2020)]%
        {zhu2020deep}
\bibfield{author}{\bibinfo{person}{Heming Zhu}, \bibinfo{person}{Yu Cao},
  \bibinfo{person}{Hang Jin}, \bibinfo{person}{Weikai Chen},
  \bibinfo{person}{Dong Du}, \bibinfo{person}{Zhangye Wang},
  \bibinfo{person}{Shuguang Cui}, {and} \bibinfo{person}{Xiaoguang Han}.}
  \bibinfo{year}{2020}\natexlab{}.
\newblock \showarticletitle{Deep {Fashion3D}: {A} dataset and benchmark for
  {3D} garment reconstruction from single images}. In
  \bibinfo{booktitle}{\emph{European Conference on Computer Vision (ECCV)}}.
  Springer, \bibinfo{pages}{512--530}.
\newblock


\bibitem[Zhu et~al\mbox{.}(2022)]%
        {zhu2022registering}
\bibfield{author}{\bibinfo{person}{Heming Zhu}, \bibinfo{person}{Lingteng Qiu},
  \bibinfo{person}{Yuda Qiu}, {and} \bibinfo{person}{Xiaoguang Han}.}
  \bibinfo{year}{2022}\natexlab{}.
\newblock \showarticletitle{Registering Explicit to Implicit: Towards
  High-Fidelity Garment mesh Reconstruction from Single Images}. In
  \bibinfo{booktitle}{\emph{Proceedings of the IEEE/CVF Conference on Computer
  Vision and Pattern Recognition}}. \bibinfo{pages}{3845--3854}.
\newblock


\end{thebibliography}

\newpage
\appendix
\section{Appendix}

The supplementary material includes this document and an additional video. Here, we provide more details about the datasets and the implementation, and present further results. 

\subsection{Implementation Details}
We choose $\sigma=0.1$, $|\mathcal{N}\left(\vect{x}\right)| = 6$, $t_n = -0.6$, and $t_f = 0.6$ and weight the individual losses with
$\lossweight{\text{vol}}=1.0$,
$\lossweight{\text{mrf}}=0.0005$,
$\lossweight{\text{clothing}}=0.5$,
$\lossweight{\text{silhouette}}=0.001$,
$\lossweight{\text{bodymask}}=30$,
$\lossweight{\text{skin}}=1.0$,
$\lossweight{\text{inside}}=40$,
$\lossweight{\text{skininside}}=0.01$,
$\lossweight{\text{lap}}=500$,
$\lossweight{\text{offset}}=400$. 
For $\lossweight{\text{offset}}$, the weight ratio of body, face and hands region is $2:3:12$. 
Note that it is important to perform the first stage \ac{NeRF} training without optimizing the non-rigid deformation model. In this stage, we also set $\lossweight{\text{mrf}}=0$. 
In the second stage, the non-rigid deformation model then explains clothing deformations that cannot be explained by the body transformation. And $L_{mrf}$ helps capture more details that can not be modelled by the non-rigid deformation. 
The overall optimization time is around 40 hours with NVIDIA V100.

\subsection{Datasets}
We use 4 subjects ('male-3-casual', 'female-3-casual', 'male-4-casual', 'female-4-casual') from People Snapshot \cite{alldieck2018video} for qualitative and quantitative evaluation. 
We follow the settings of Anim-NeRF \cite{chen2021animatable}, namely
\begin{itemize}
    \item 'male-3-casual': frames 1-456 with step size of 4 for training, and frames 456-676 with step size 4 for test.
    \item 'male-4-casual': frame 1-660 with step size 6 for training, and frames 661-873 with step size 4 for test.
    \item 'female-3-casual': frame 1-446 with step size 4 for training, and frames 447-648 with step size 4 for test.
    \item 'female-4-casual': frame 1-336 with step size 4 for training, and frames 336-524 with step size 4 for test.
\end{itemize}
We further use 4 subjects ('subject003', 'subject016', 'subject022', 'subject023') with outfit 1 and motion 1 from iPER \cite{lwb2019} for qualitative evaluation.  
For all subjects, we use frames 1-490 with step 4 for optimization. 
We use 4 synthetic video data ('female outfit1', 'female outfit2', 'female outfit3', 'male outfit1') and 1 self-captured video ('CHH female') from SelfRecon \cite{jiang2022selfrecon}. For each subject, we use 100 frames for optimization. 
For self-captured data, we record videos of each subject wearing different clothing types. The subject wears different clothes and performs an A-pose video and a video with random actions. In our experiments, we use A-pose videos of subject 'Yao' with six types of clothing for qualitative evaluation, those videos include loose dressing and short skirts. For each video, we use frames 0-400 with step 2 for optimization. 
\subsection{Ablation Experiments}

\begin{figure}[h]
    \includegraphics[width=\linewidth]{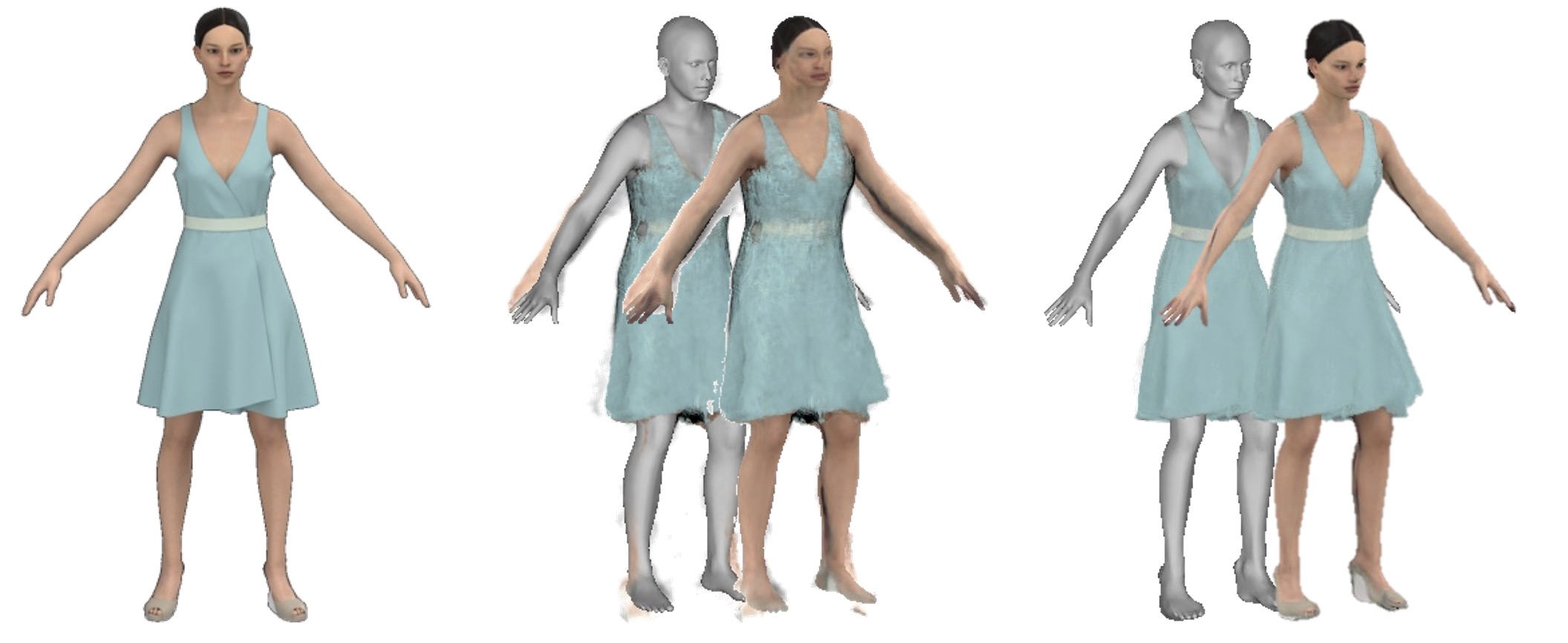}\\
    \small{Source subject \hspace{1.8em} w/o pose refinement \hspace{2.0em} w/ pose refinement}
    \vspace{-0.8em}
	\caption{Novel view synthesis w/o and w/ pose refinement. The pose refinement improves the visual quality of the reconstruction, as more texture details are reconstructed.}
	\label{fig:supp_ablation_pose}
	\vspace{-0.8em}
\end{figure}

\qheading{Effect of pose refinement}
Since the pose estimation for each frame is not accurate, the pose refinement is important to gain details. 
We try learning our method without pose refinement. 
Fig. \ref{fig:supp_ablation_pose} shows that pose refinement improves the image quality a lot. 

\subsection{More Qualitative results}

\begin{figure}[h]
    \includegraphics[width=\linewidth]{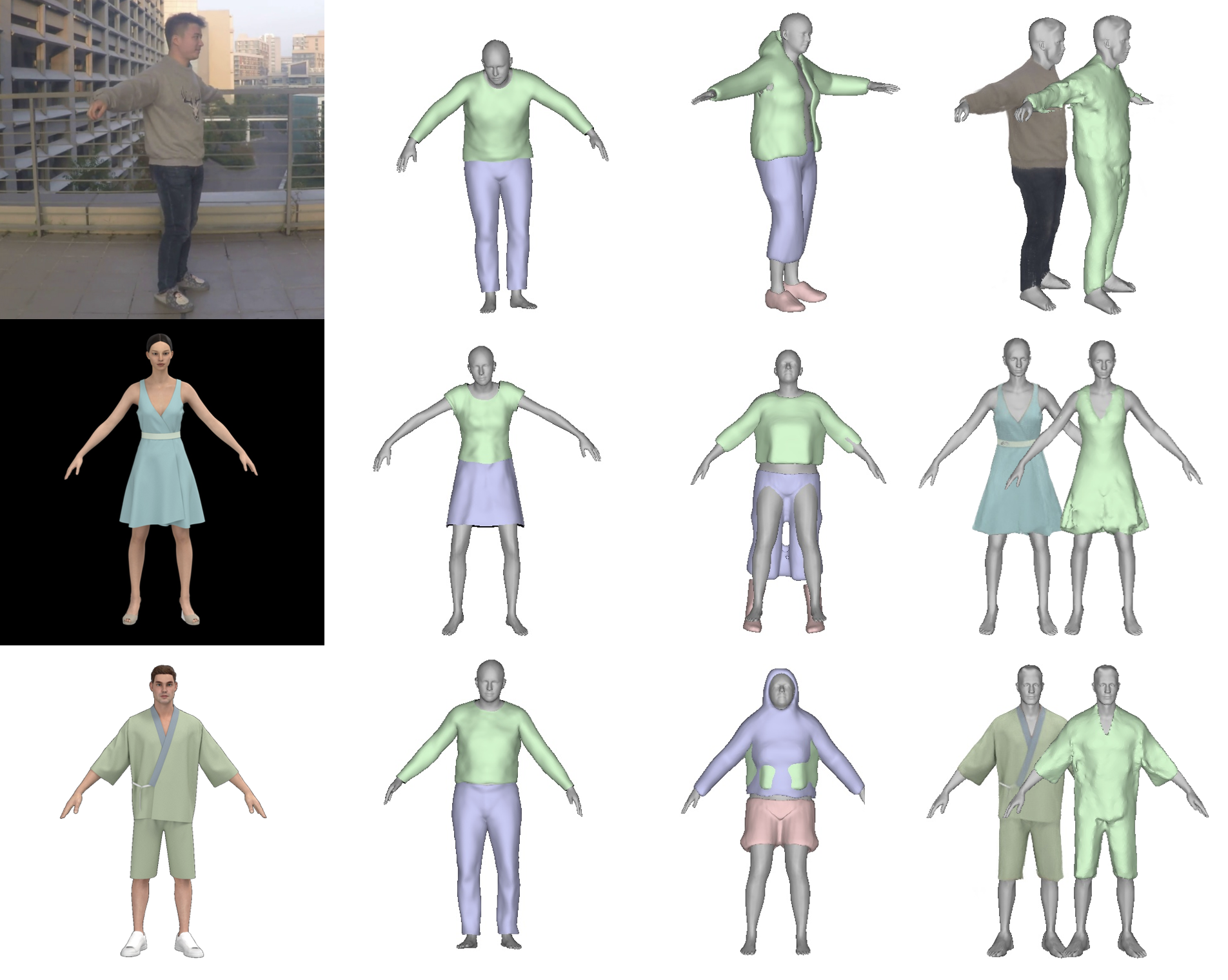}\\
    \small{Input image \hspace{1.5em} SMPLicit \hspace{3em} BCNet \hspace{5em} Ours \hspace{2em}}
    \vspace{-0.8em}
	\caption{Additional examples for qualitative comparison of garment reconstruction. \modelname reconstructs different clothing types more faithfully than SMPLicit \cite{corona2021smplicit} and BCNet \cite{jiang2020bcnet}.}
	\label{fig:comparison_garment_supp}
	\vspace{-0.8em}
\end{figure}

We show additional comparisons on Garment reconstruction with SMPLicit \cite{corona2021smplicit} and BCNet \cite{jiang2020bcnet} in Fig~\ref{fig:comparison_garment_supp}.
\modelname gives better visual quality than SMPLicit and BCNet. 
Note that the training/optimization settings are different, they reconstruct the body and garment from a single image, while our results are learned from video. 
However, they require a large set of 3D scans and manually designed cloth template for training, while we do not need any 3D supervision, and capture the garment appearance as well. 

\begin{figure}[h]
    \includegraphics[width=\linewidth]{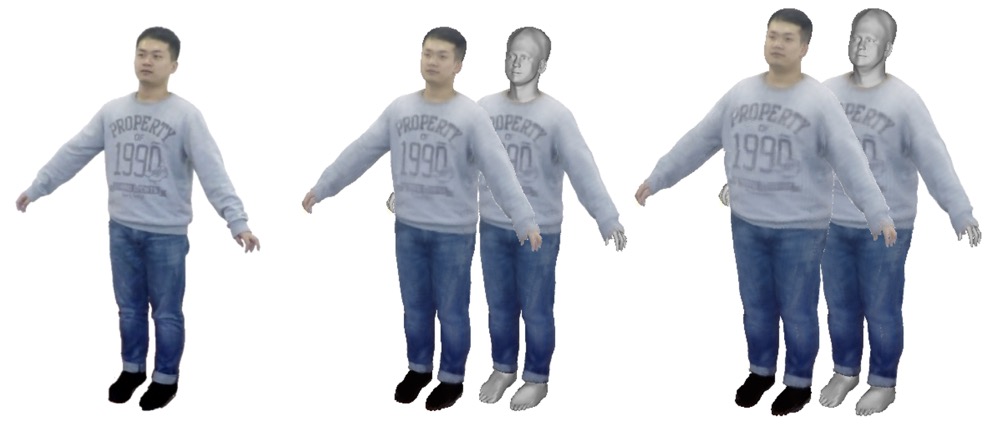}\\
    \small{Source subject \hspace{3.5em} Reconstruction \hspace{4em} Changing shape}
    \vspace{-0.8em}
    \caption{SCARF can change underlying body shapes by altering SMPL-X shape parameters, the NeRF clothing will adapt to the body accordingly.}
    \vspace{-0.5em}
    \label{fig:alter_shape}
\end{figure}

In Figure~\ref{fig:alter_shape}, we also show that SCARF can alter body shape and the clothing will adapt to the shape accordingly.

\subsection{Limitations}
\begin{figure}[h]
    \includegraphics[width=\linewidth]{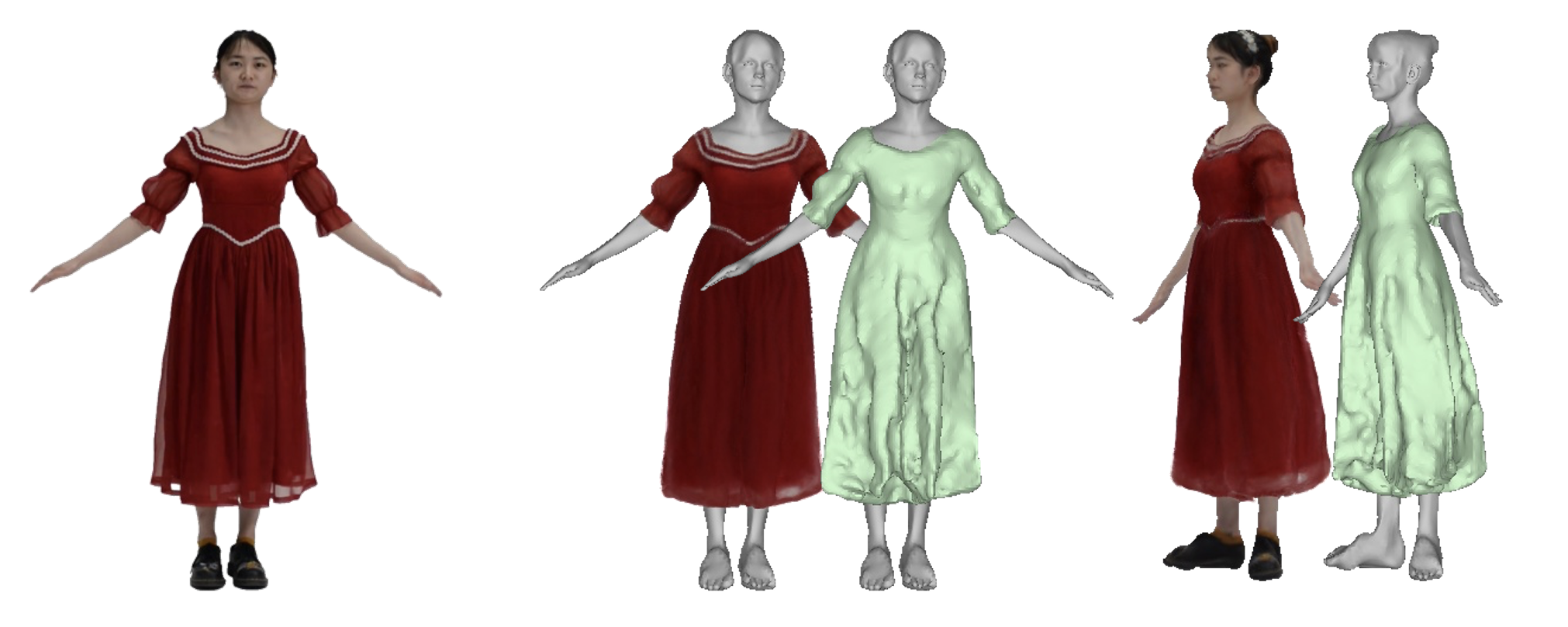}\\
    \small{Reference image \hspace{4em} Captured clothing appearance \& geometry \hspace{5em} }
    \vspace{-0.8em}
	\caption{While \modelname gives good visual quality for clothing renderings, the underlying geometry of the \ac{NeRF} clothing is sometimes noisy.}
	\label{fig:limitation_geometry}
	\vspace{-0.8em}
\end{figure}

\begin{figure}[htbp]
    \includegraphics[width=\linewidth]{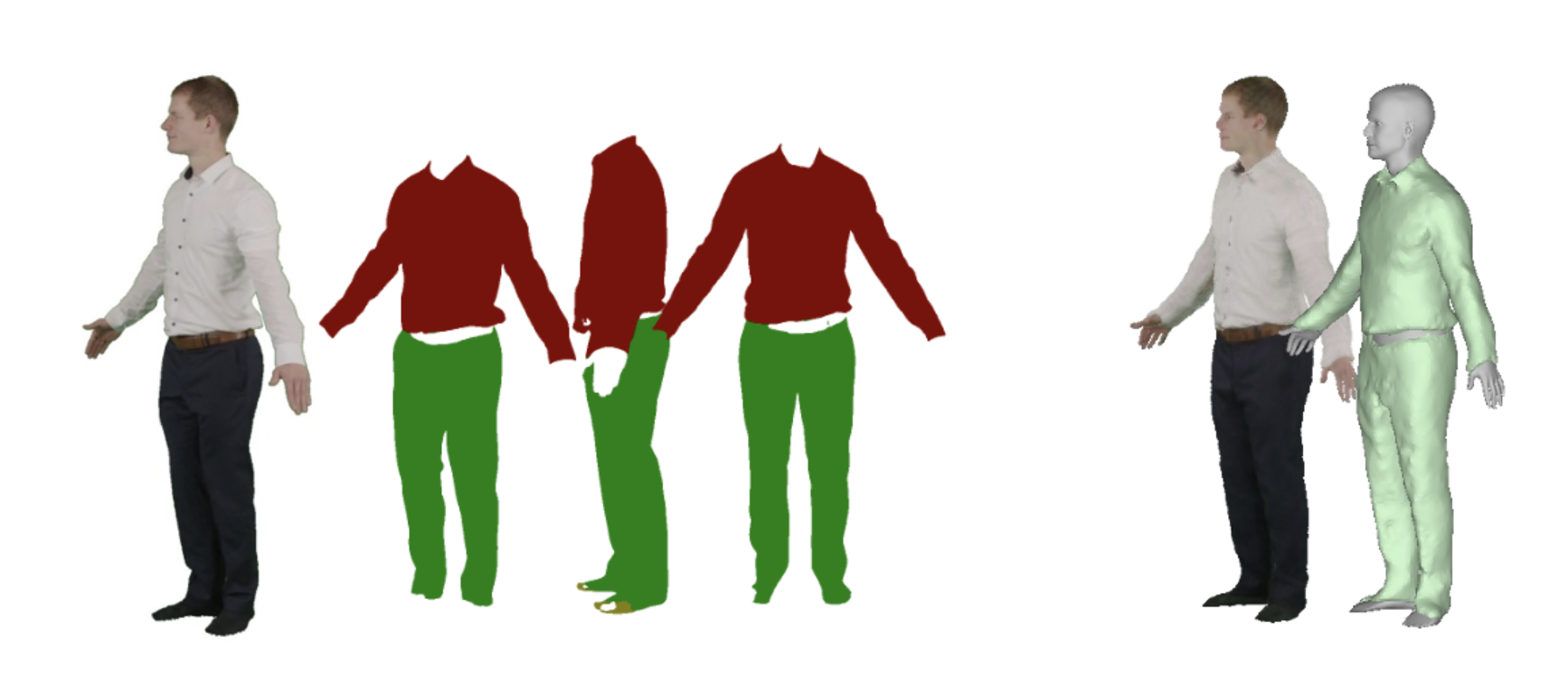}\\
    \small{\hspace{1em} Input image and clothing segmentations \hspace{4.5em} Reconstruction}
    \vspace{-0.8em}
	\caption{The wrong clothing segmentation results in a visible gap within the reconstructed clothing.}
	\label{fig:limitation_segmentation}
	\vspace{-0.8em}
\end{figure}

\begin{figure}[h]
    \begin{tabular}{c@{}c@{}c}
        \includegraphics[width=0.33\columnwidth]{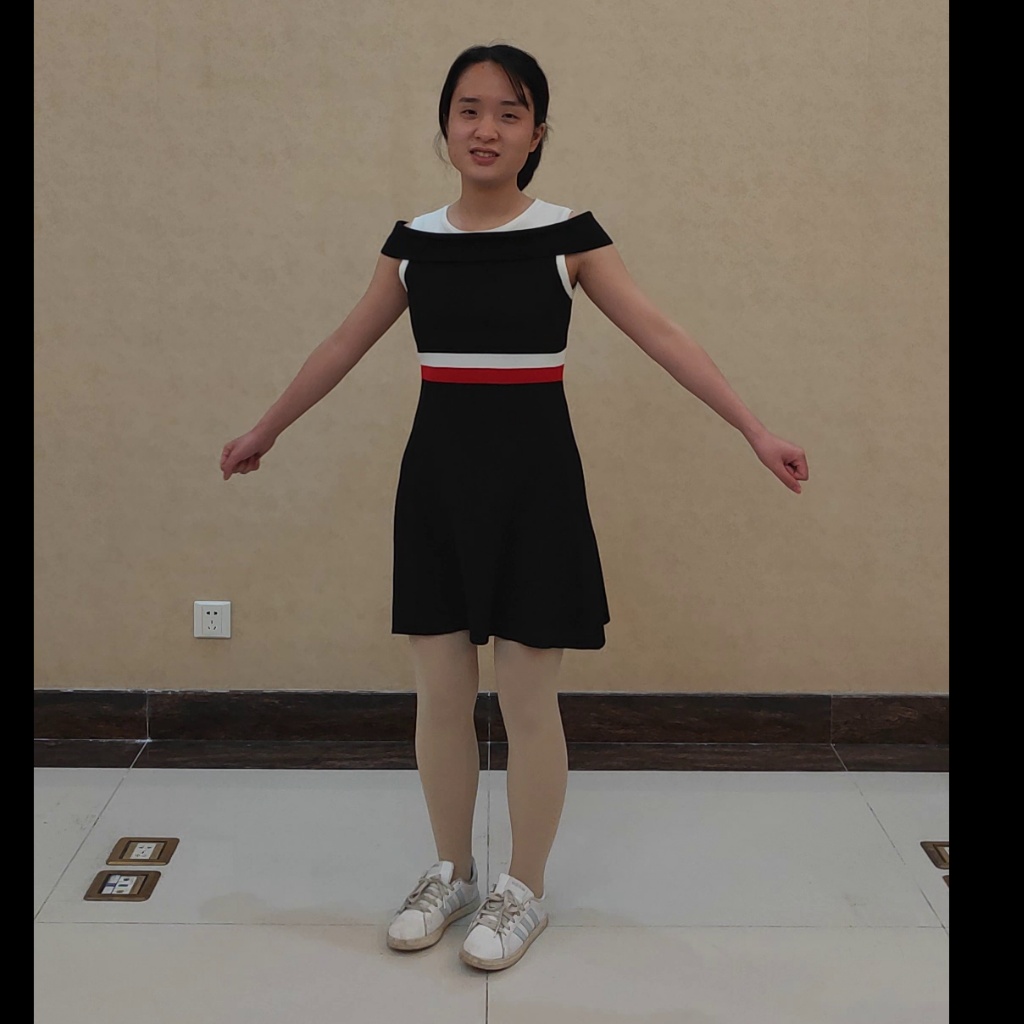} &  
        \includegraphics[width=0.33\columnwidth]{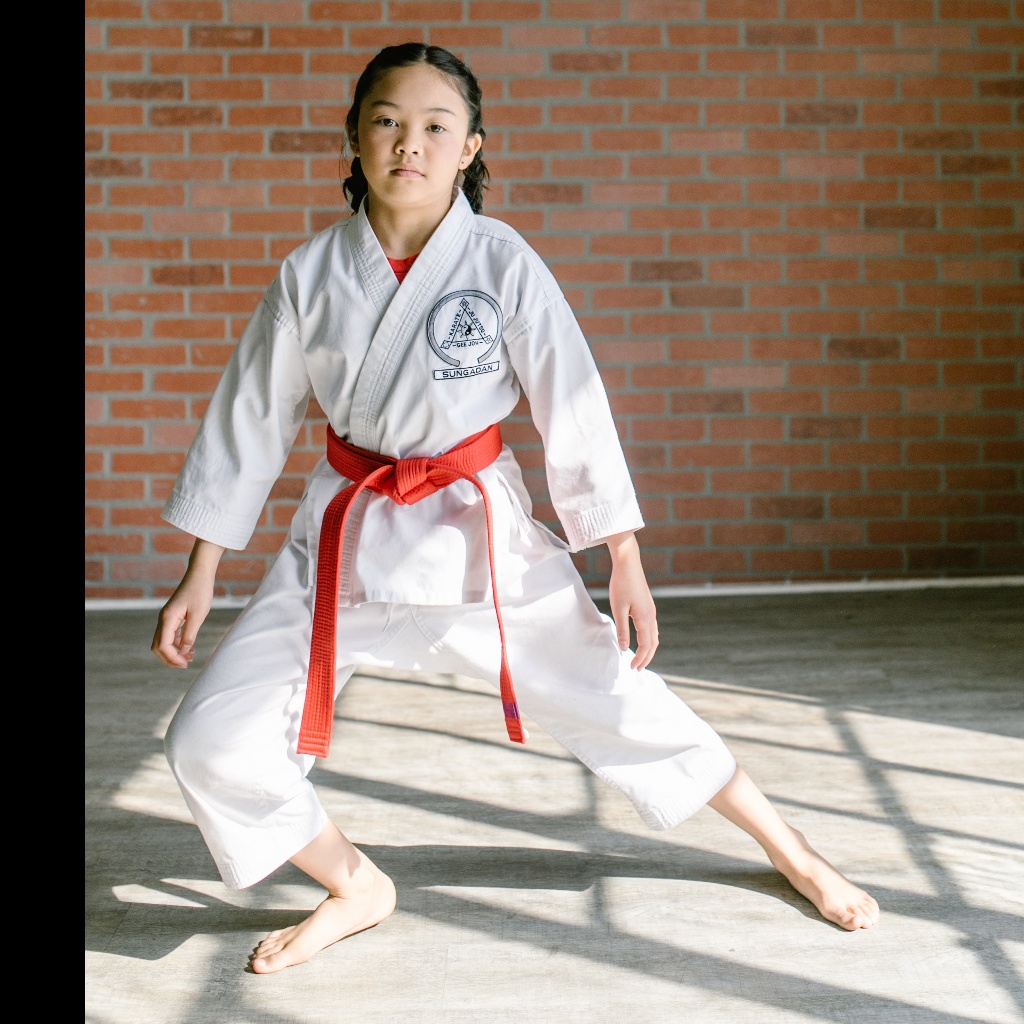} & 
        \includegraphics[width=0.33\columnwidth]{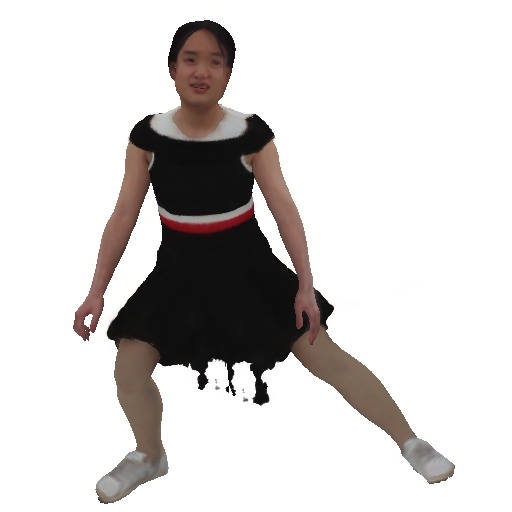}  \\
        \includegraphics[width=0.33\columnwidth]{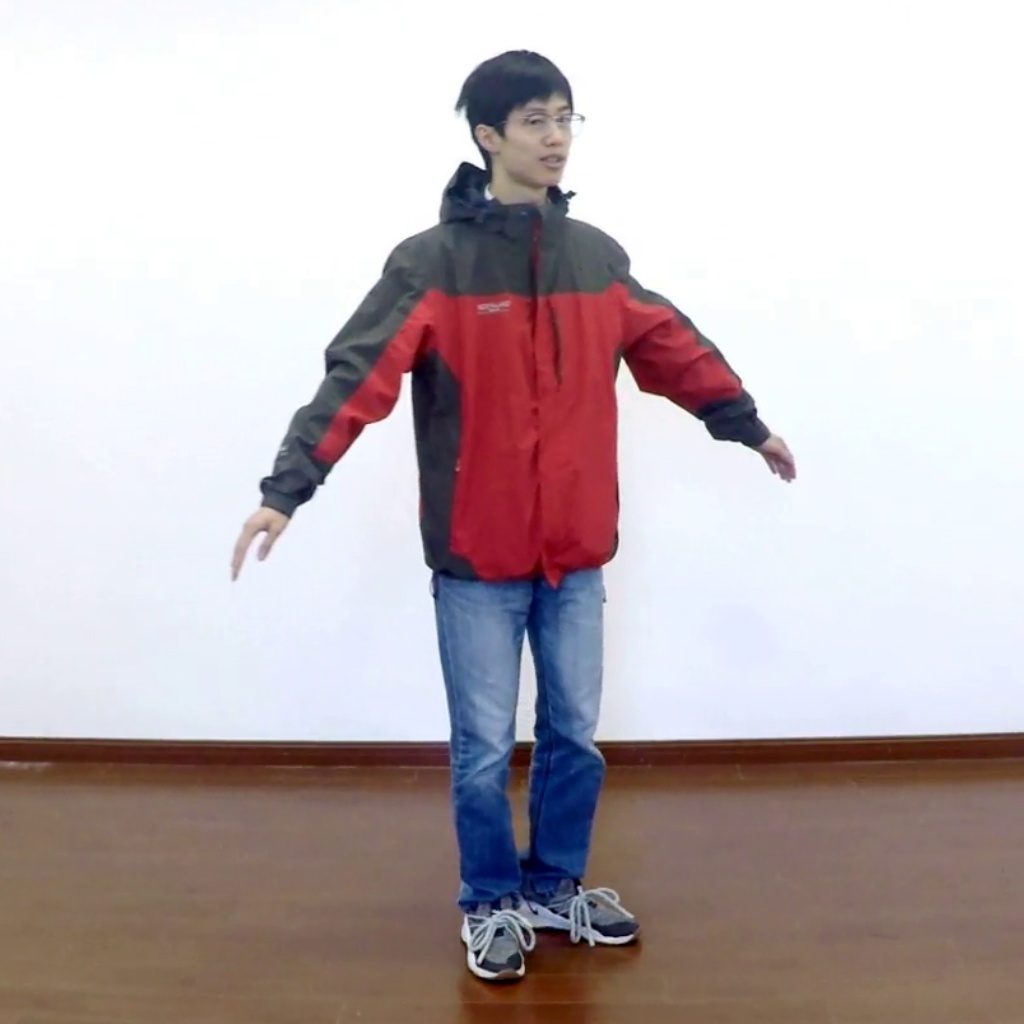} &  
        \includegraphics[width=0.33\columnwidth]{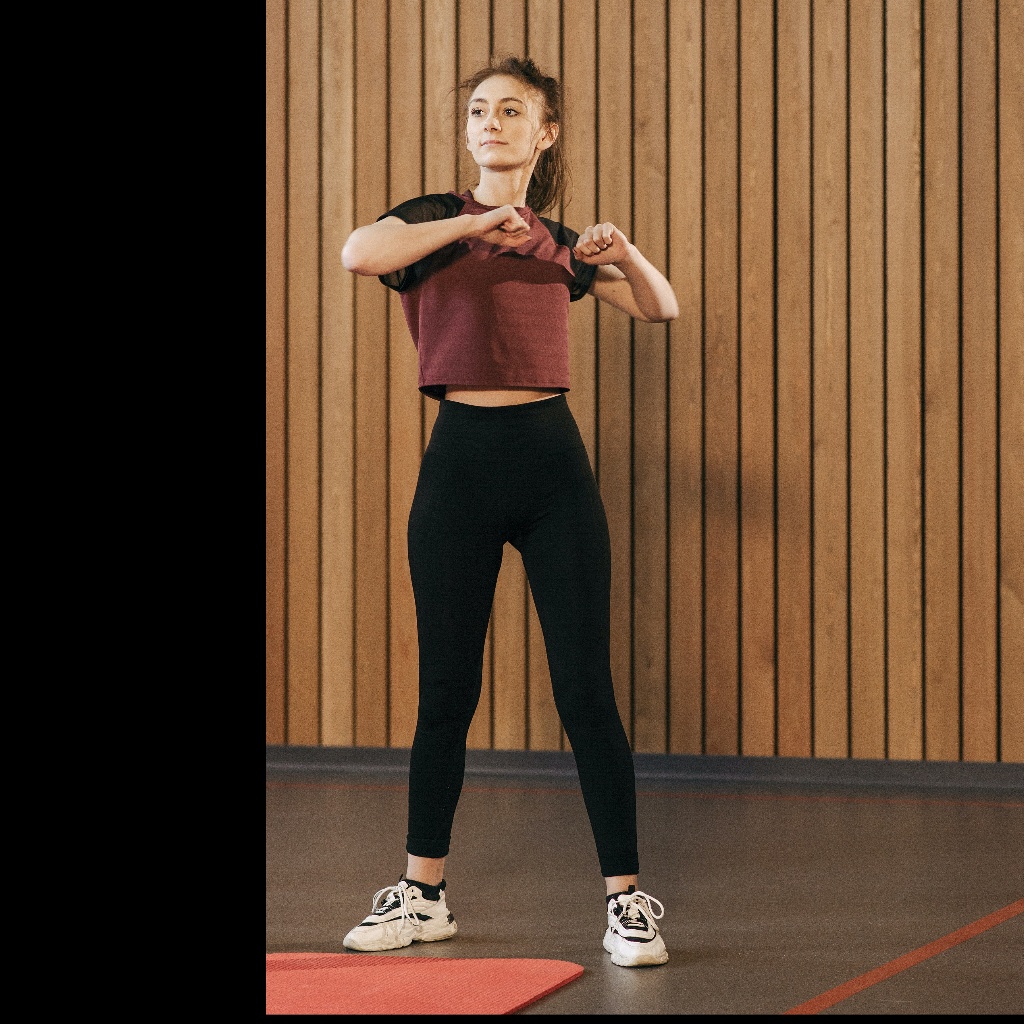} & 
        \includegraphics[width=0.33\columnwidth]{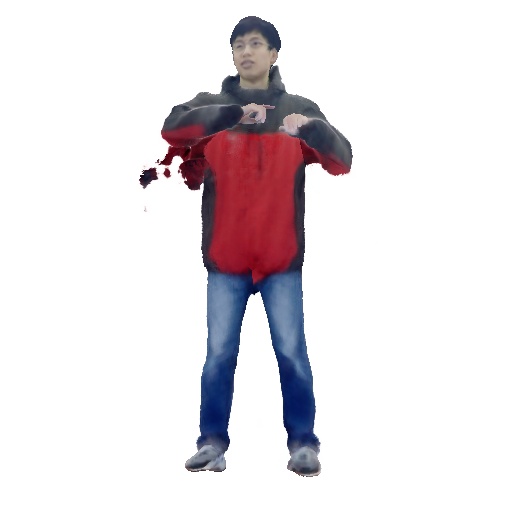}  \\
         Input subject &
         Reference pose &
         Animation 
    \end{tabular}
	\caption{Reposing can result in visual artifacts for unseen poses.}
	\label{fig:limitation_animation}
\end{figure}

The main paper discusses limitations regarding the image segmentation, the quality of the reconstructed geometry, and the generalization to unseen poses. 
Figure~\ref{fig:limitation_segmentation} shows the wrong reconstruction due to consistent clothing segmentation errors, e.g. the belt is not recognized as part of clothing in segmentation, this results in wrong disentanglement between human body and clothing. 
Fig. \ref{fig:limitation_geometry} shows an example of noisy geometry despite good visual quality, 
and Figure~\ref{fig:limitation_animation} shows some reposing artifacts for unseen poses. 

\end{document}